\documentclass[]{article} 
\usepackage{proceed2e}

\usepackage{graphicx}
\usepackage{times}
\usepackage{amsmath}
\usepackage{amssymb}
\usepackage{paralist}
\usepackage{color}
\usepackage[round,sort&compress]{natbib}
\usepackage{url}

\DeclareMathOperator{\assign}{:\!=}

\usepackage{xspace}
\def\onedot{\ifx\@let@token.\else.\null\fi\xspace}
\def\eg{{e.g}\onedot}

\def\etal{{et al}\onedot}

\begin{document}

\title{Simultaneous Object Detection, Tracking, and Event Recognition}

\author{
  Andrei Barbu, Aaron Michaux, Siddharth Narayanaswamy, and Jeffrey Mark Siskind}
\date{Purdue University\\
School of Electrical and Computer Engineering\\
465 Northwester Avenue\\
West Lafayette IN 47907-2035 USA\\
{\tt\small andrei@0xab.com,\{amichaux,snarayan,qobi\}@purdue.edu}}

\maketitle

\let\thefootnote\relax\footnotetext{Additional images and videos as well as all
  code and datasets are available at
  \url{http://engineering.purdue.edu/~qobi/arxiv2012a}.}

\begin{abstract}
  The common internal structure and algorithmic organization of object
  detection, detection-based tracking, and event recognition facilitates a
  general approach to integrating these three components.
  This supports multidirectional information flow between these components
  allowing object detection to influence tracking and event recognition and
  event recognition to influence tracking and object detection.
  The performance of the combination can exceed the performance of the
  components in isolation.
  This can be done with linear asymptotic complexity.
\end{abstract}

\section{Introduction}

Many common approaches to event recognition \citep{Siskind1996, Starner98,
  Wang2009, Xu2002, Xu2005} classify events based on their motion profile.
This requires detecting and tracking the event participants.
Adaptive approaches to tracking \citep{Yilmaz2006}, \eg\ Kalman filtering
\citep{Comaniciu2003}, suffer from three difficulties that impact their utility
for event recognition.
First, they must be initialized.
One cannot initialize on the basis of motion since many event participants move
only for a portion of the event, and sometimes not at all.
Second, they exhibit drift and often must be periodically reinitialized to
compensate.
Third, they have difficulty tracking small, deformable, or partially
occluded objects as well as ones whose appearance changes dramatically.
This is particularly of concern since many events, \eg\ picking things up,
involve humans interacting with objects that are sufficiently small for humans
to grasp and where such interaction causes appearance change by out-of-plane
rotation, occlusion, or deformation.

Detection-based tracking is an alternate approach that attempts to address
these issues.
In detection-based tracking an object detector is applied to each frame of a
video to yield a set of candidate detections which are composed into tracks by
selecting a single candidate detection from each frame that maximizes temporal
coherency of the track.
However, current object detectors are far from perfect.
On the PASCAL VOC Challenge, they typically achieve average precision scores
of 40\% to 50\% \citep{Everingham10}.
Directly applying such detectors on a per-frame basis would be ill-suited
to event recognition.
Since the failure modes include both false positives and false negatives,
interpolation does not suffice to address this shortcoming.
A better approach is to combine object detection and tracking with a single
objective function that maximizes temporal coherency to allow object detection
to inform the tracker and vice versa.

One can carry this approach even further and integrate event recognition with
both object detection and tracking.
One way to do this is to incorporate coherence with a target event model into
the temporal coherency measure.
For example, a top-down expectation of observing a \emph{pick up} event
can bias the object detector and tracker to search for event participants
that exhibit the particular joint motion profile of that event: an object in
close proximity to the agent, the object starting out at rest while the agent
approaches the object, then the agent touching the object, followed by the
object moving with the agent.
Such information can also flow bidirectionally.
Mutual detection of a \emph{baseball bat} and a \emph{hitting} event can be
easier than detecting each in isolation or having a fixed direction of
information flow.

The common internal structure and algorithmic organization of current
object detectors \citep{Felzenszwalb2010b, Felzenszwalb2010a},
detection-based trackers \citep{Wolf1989}, and HMM-based approaches to
event recognition \citep{Baum1966} facilitates a general approach to
integrating these three components.
We demonstrate an approach to integrating object detection, tracking, and event
recognition and show how it improves each of the these three components in
isolation.
Further, while prior detection-based trackers exhibit quadratic complexity, we
show how such integration can be fast, with linear asymptotic complexity.

\section{Detection-based tracking}
\label{sec:detectionbasedtracking}

The methods described in sections~\ref{sec:detectionandtracking},
\ref{sec:trackingandevents}, and \ref{sec:detectionandtrackingandevents}
extend a popular dynamic-programming approach to detection-based tracking.
We review that approach here to set forth the concepts, terminology, and
notation that will be needed to describe the extensions.

\begin{figure}
  \begin{center}
    \includegraphics[scale=0.57]{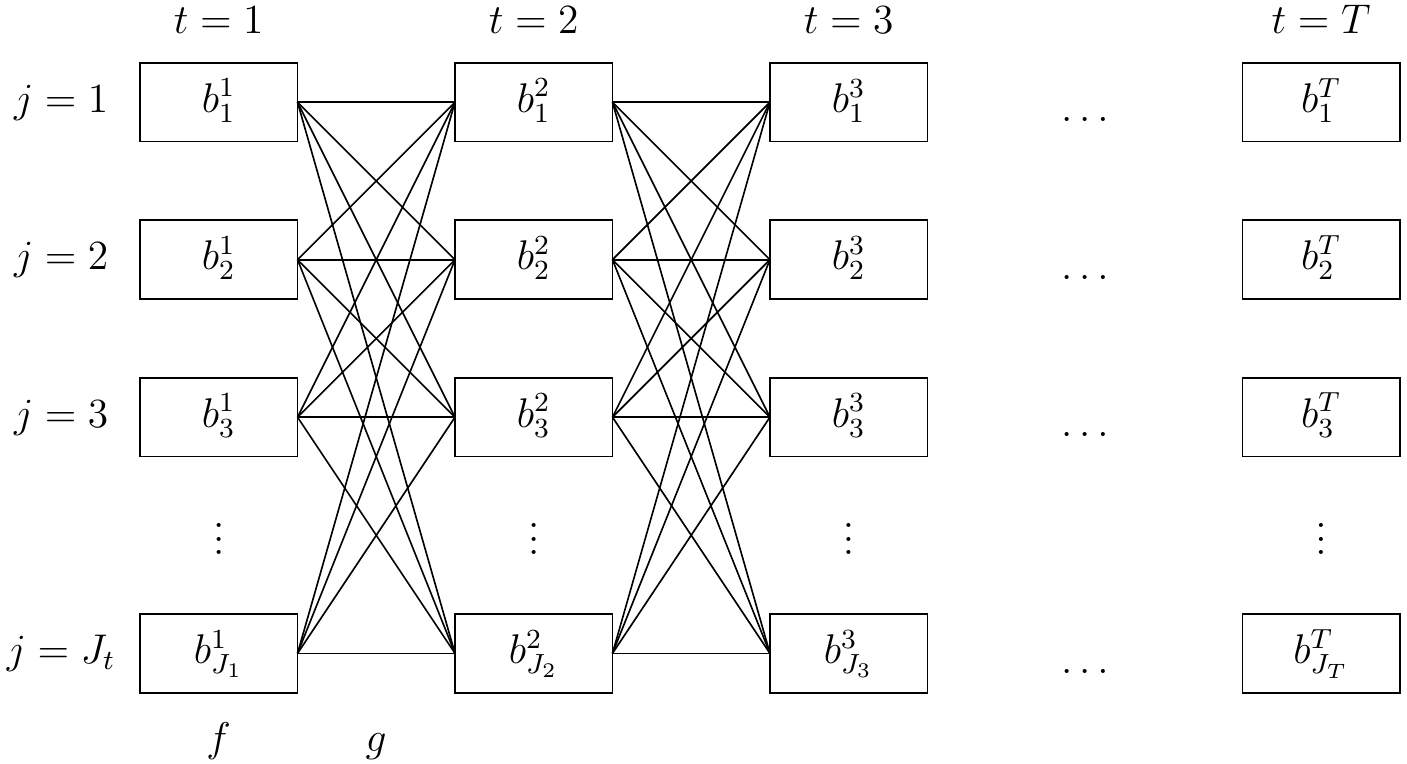}
  \end{center}
  \caption{The tracking lattice constructed by the Viterbi algorithm performing
    detection-based tracking.}
  \label{fig:viterbi}
\end{figure}

Detection-based tracking is a general framework where an object detector is
applied to each frame of a video to yield a set of candidate detections which
are composed into tracks by selecting a single candidate detection from each
frame that maximizes temporal coherency of the track.
This general framework can be instantiated with answers to the following
questions:
\begin{compactenum}
  \item What is the representation of a \emph{detection}?
    \label{a}
  \item What is the \emph{detection source}?
    \label{b}
  \item What is the measure of temporal coherency?
    \label{c}
  \item What is the procedure for finding the track with maximal temporal
    coherency?
    \label{d}
\end{compactenum}
We answer questions~\ref{a} and~\ref{b} by taking a detection to be a
scored axis-aligned rectangle (box), such as produced by the
\cite{Felzenszwalb2010b, Felzenszwalb2010a} object detectors, though
our approach is compatible with any method for producing scored
axis-aligned rectangular detections.
If~$b^t_j$ denotes the~$j$th detection in frame~$t$, $f(b^t_j)$ denotes the
score of that detection, $T$~denotes the number of frames, and
$\mathbf{j}=\langle j_1,\ldots,j_T\rangle$ denotes a track comprising
the~$j_t$th detection in frame~$t$, we answer question~\ref{c} by formulating
temporal coherency of a track $\mathbf{j}=\langle j_1,\ldots,j_T\rangle$ as:
\begin{equation}
  \max_{j_1,\ldots,j_T}
  \sum_{t=1}^T f(b^t_{j_t})+
  \sum_{t=2}^T g(b^{t-1}_{j_{t-1}},b^t_{j_t})
\label{eq:track}
\end{equation}
where~$g$ scores the local temporal coherency between detections in adjacent
frames.
We take~$g$ to be the negative Euclidean distance between the center of
$b^t_{j_t}$ and the center of $b^{t-1}_{j_{t-1}}$ projected forward one frame,
though, as discussed below, our approach is compatible with a variety of
functions discussed by \cite{Felzenszwalb2004}.
The forward projection internal to~$g$ can be done in a variety of ways
including optical flow and the Kanade-Lucas-Tomasi (KLT)
\citep{shi1994, tomasi1991} feature tracker.
We answer question~\ref{d} by observing that Eq.~\ref{eq:track} can be
optimized in polynomial time with the Viterbi algorithm \citep{Viterbi1971}:
\begin{equation}
  \begin{array}[t]{@{}l@{}}
    \textbf{for}\;j=1\;\textbf{to}\;J_1\;
    \textbf{do}\;\delta^1_j\assign f(b^1_j)\\
    \textbf{for}\;t=2\;\textbf{to}\;T\\
    \textbf{do}\;\begin{array}[t]{@{}l@{}}
		 \textbf{for}\;j=1\;\textbf{to}\;J_t\\
		 \textbf{do}\;\delta^t_j\assign
			       f(b^t_j)
			       +\displaystyle
				\max_{j'=1}^{J_{t-1}} g(b^{t-1}_{j'},b^t_j)
						   +\delta^{t-1}_{j'}
		 \end{array}
  \end{array}
  \label{eq:viterbi}
\end{equation}
\noindent
where~$J_t$ is the number of detections in frame~$t$.
This leads to a lattice as shown in Fig.~\ref{fig:viterbi}.

Detection-based trackers exhibit less drift than adaptive approaches to tracking
due to fixed target models.
They also tend to perform better than simply picking the best detection in each
frame.
The reason is that one can allow the detection source to produce multiple
candidates and use the combination of the detection score~$f$ and the
adjacent-frame temporal-coherency score~$g$ to select the track.
The essential attribute of detection-based tracking is that~$g$ can
overpower~$f$ to assemble a more coherent track out of weaker detections.
The nonlocal nature of Eq.~\ref{eq:track} can allow more-reliable tracking
with less-reliable detection sources.

A crucial practical issue arises: \emph{How many candidate detections should be
produced in each frame?}
Producing too few may risk failing to produce the desired detection that is
necessary to yield a coherent track.
In the limit, it is impossible to construct any track if even a single frame
lacks any detections.\footnote{One can ameliorate this somewhat by constructing
a lattice that skips frames \citep{Sala2010}.
This increases the asymptotic complexity to be exponential in the number of
frame skips allowed.}
The current state-of-the-art in object detection is unable to simultaneously
achieve high precision and recall and thus it is necessary to explore the
trade-off between the two \citep{Everingham10}.
A detection-based tracker can bias the detection source to yield higher recall
at the expense of lower precision and rely on temporal coherency to compensate
for the resulting lower precision.
This can be done in at least three ways.
First, one can depress the detection-source acceptance thresholds.
One way this can be done with the Felzenszwalb \etal\ detectors is to lower the
trained model thresholds.
Second, one can pool the detections output by multiple detection sources with
complementary failure modes.
One way this can be done is by training multiple models for people in
different poses.
Third, one can use adaptive-tracking methods to project detections forward to
augment the raw detector output and compensate for detection failure in
subsequent frames.
This can be done in a variety of ways including optical flow and KLT.\@
The essence of our paper is a more principled collection of approaches for
compensating for low recall in the object detector.

A practical issue arises when pooling the detections output by multiple
detection sources.
It is necessary to normalize the detection scores for such pooled detections
by a per-model offset.
One can derive an offset by computing a histogram of scores of the top
detection in each frame of a video and taking the offset to be the minimum of
the value that maximizes the between-class variance \citep{Otsu1979} when
bipartitioning this histogram and the trained acceptance threshold offset by a
small but fixed amount.

\begin{figure*}[t]
  \begin{center}
    \begin{tabular}{@{}c@{\hspace*{10pt}}c@{\hspace*{2pt}}c@{\hspace*{2pt}}c@{}}
      \raisebox{15pt}{(a)}&
      \includegraphics[width=0.3\textwidth]{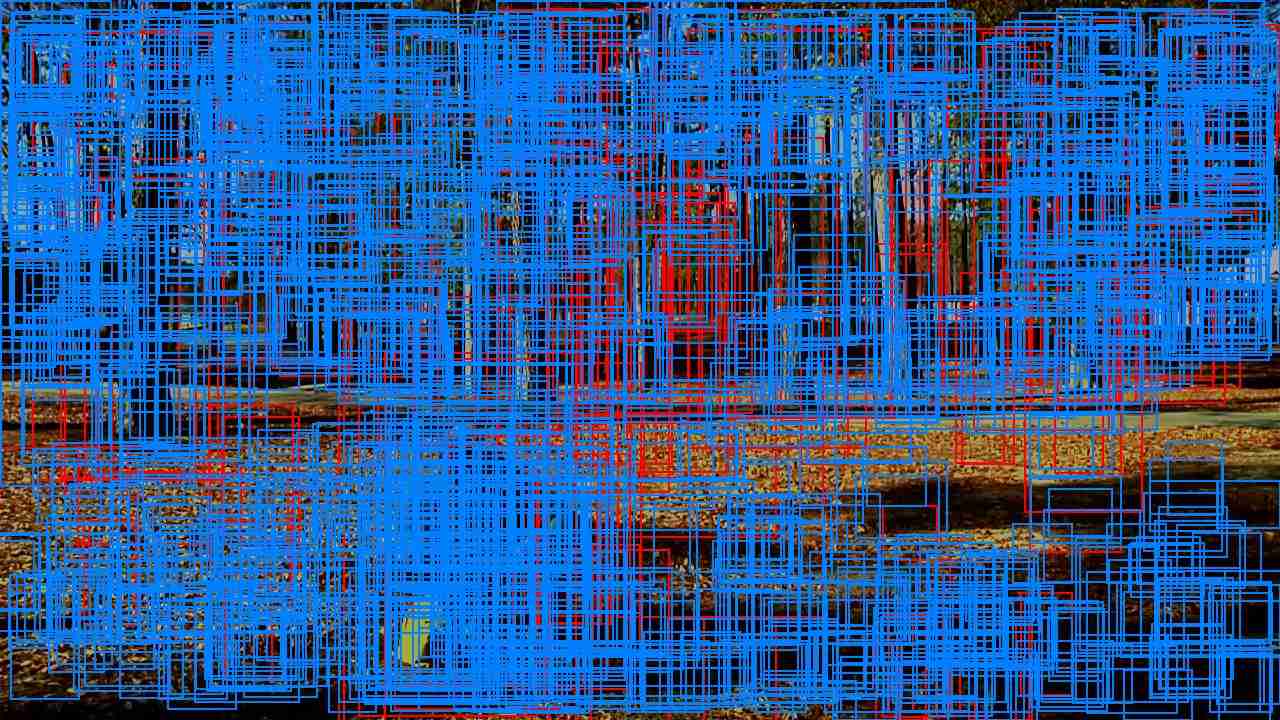}&
      \includegraphics[width=0.3\textwidth]{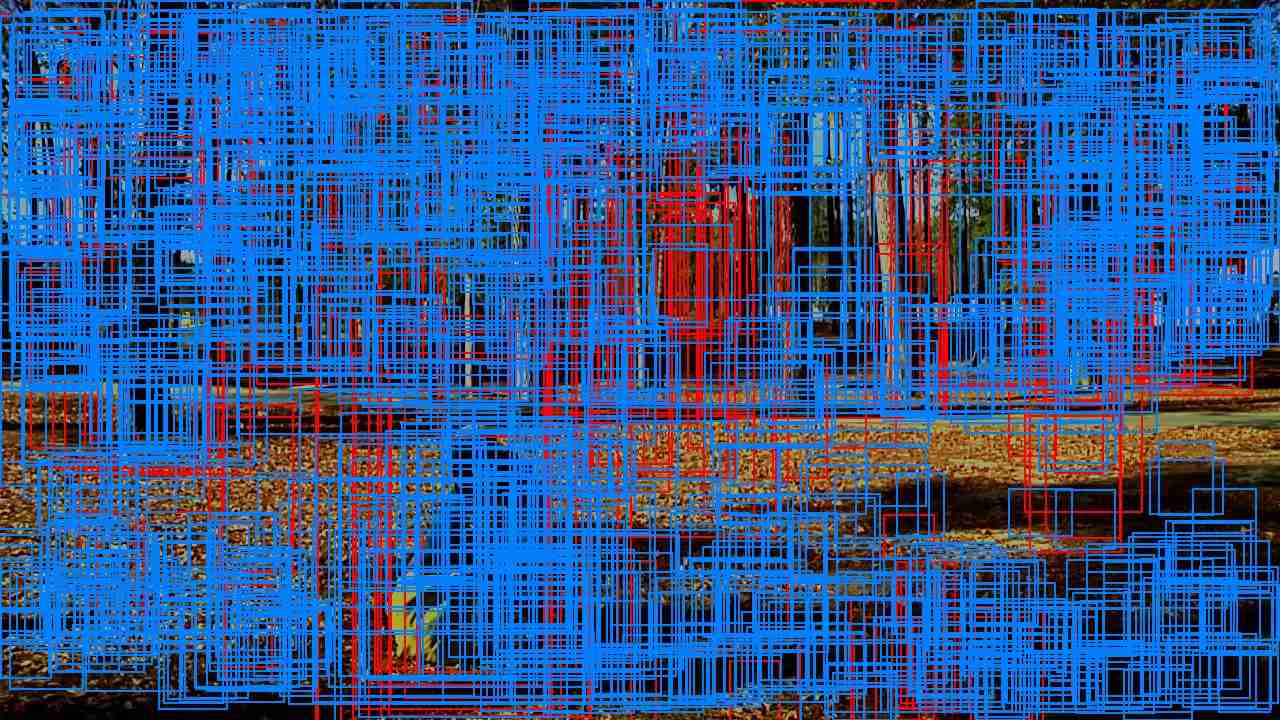}&
      \includegraphics[width=0.3\textwidth]{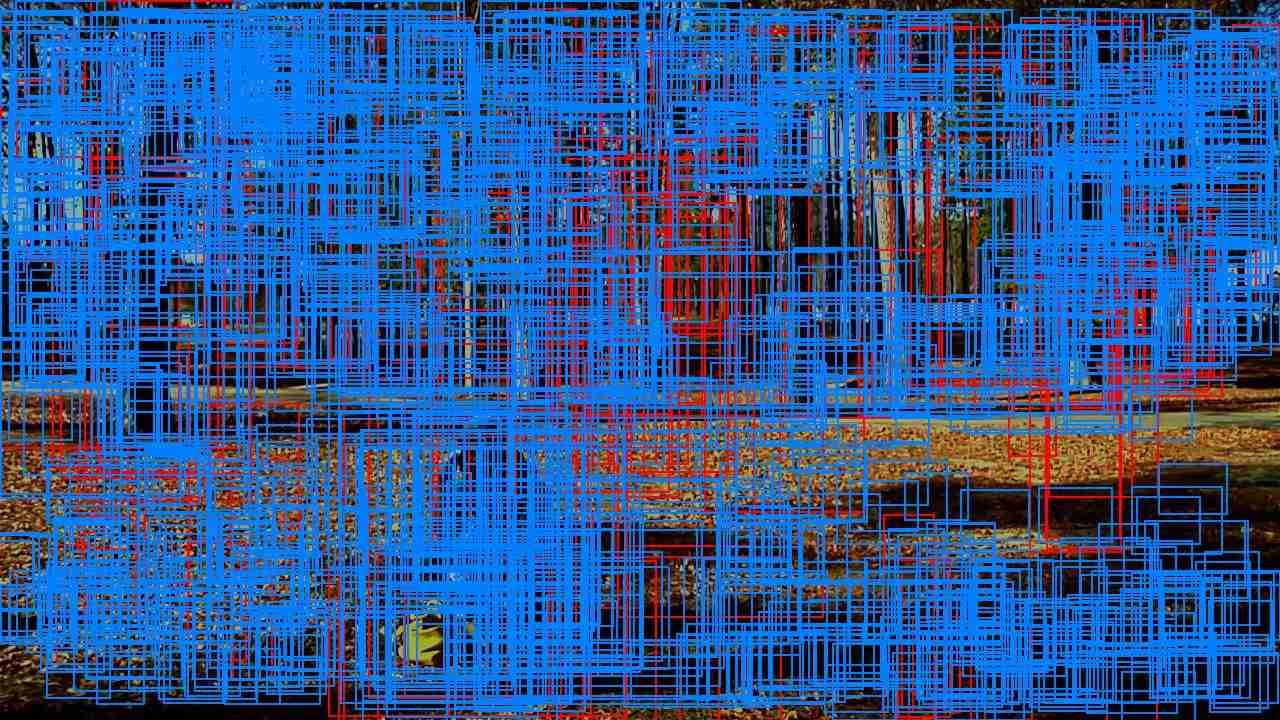}\\
      \raisebox{15pt}{(b)}&
      \includegraphics[width=0.3\textwidth]{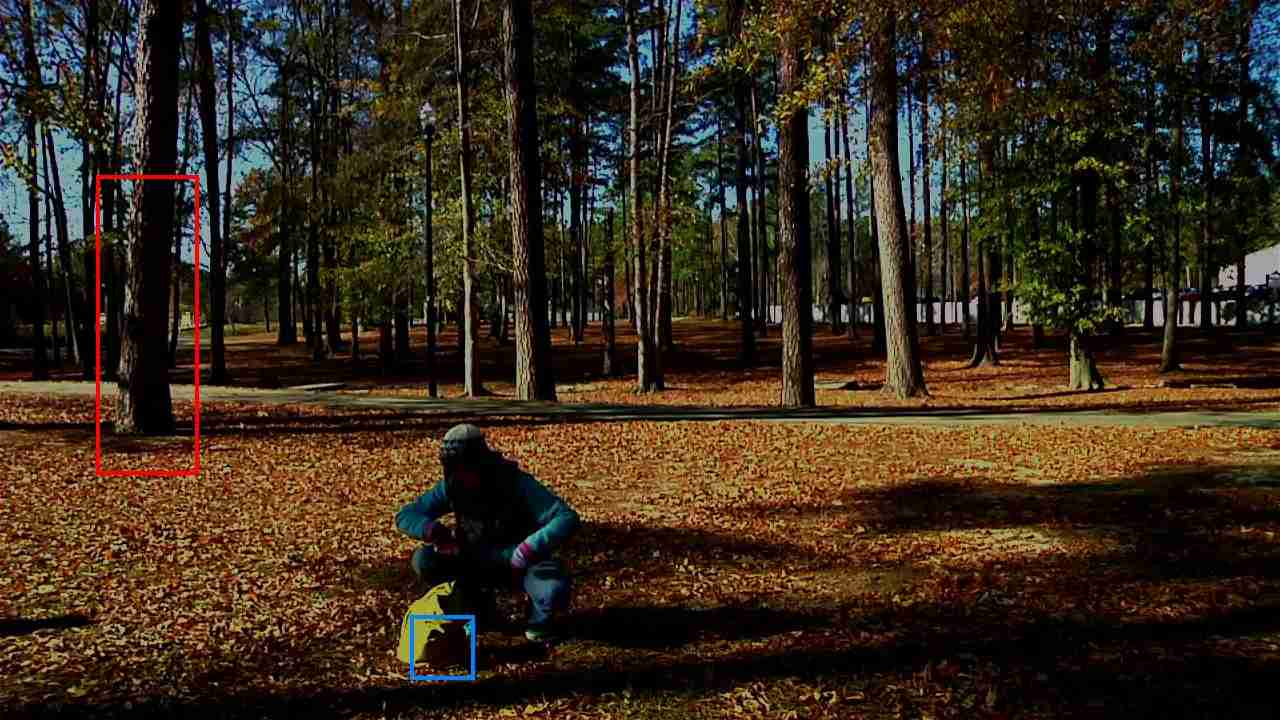}&
      \includegraphics[width=0.3\textwidth]{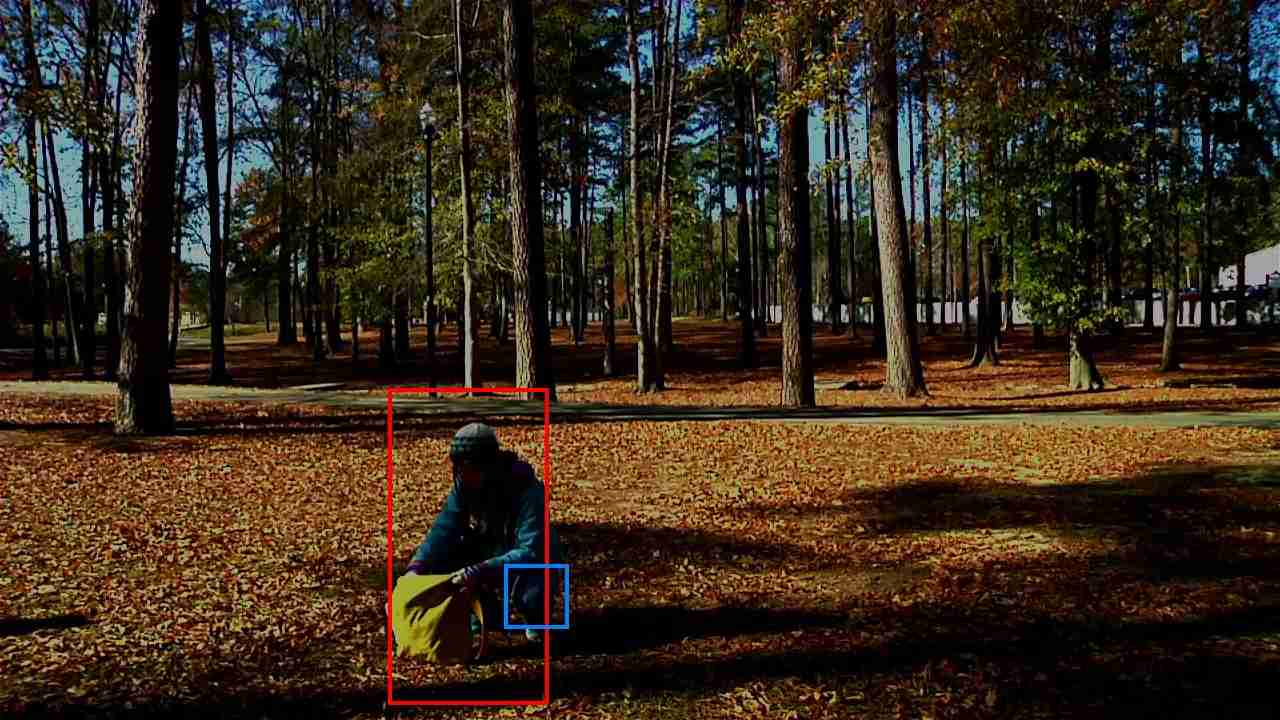}&
      \includegraphics[width=0.3\textwidth]{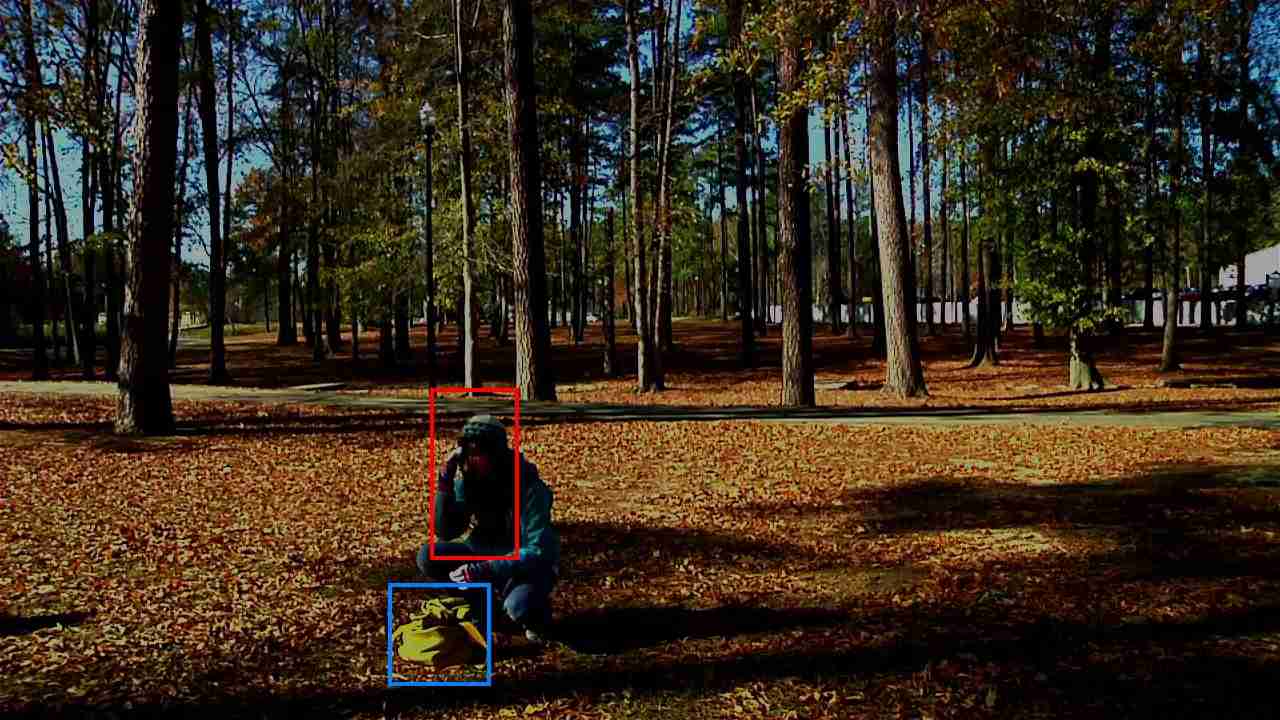}\\
      \raisebox{15pt}{(c)}&
      \includegraphics[width=0.3\textwidth]{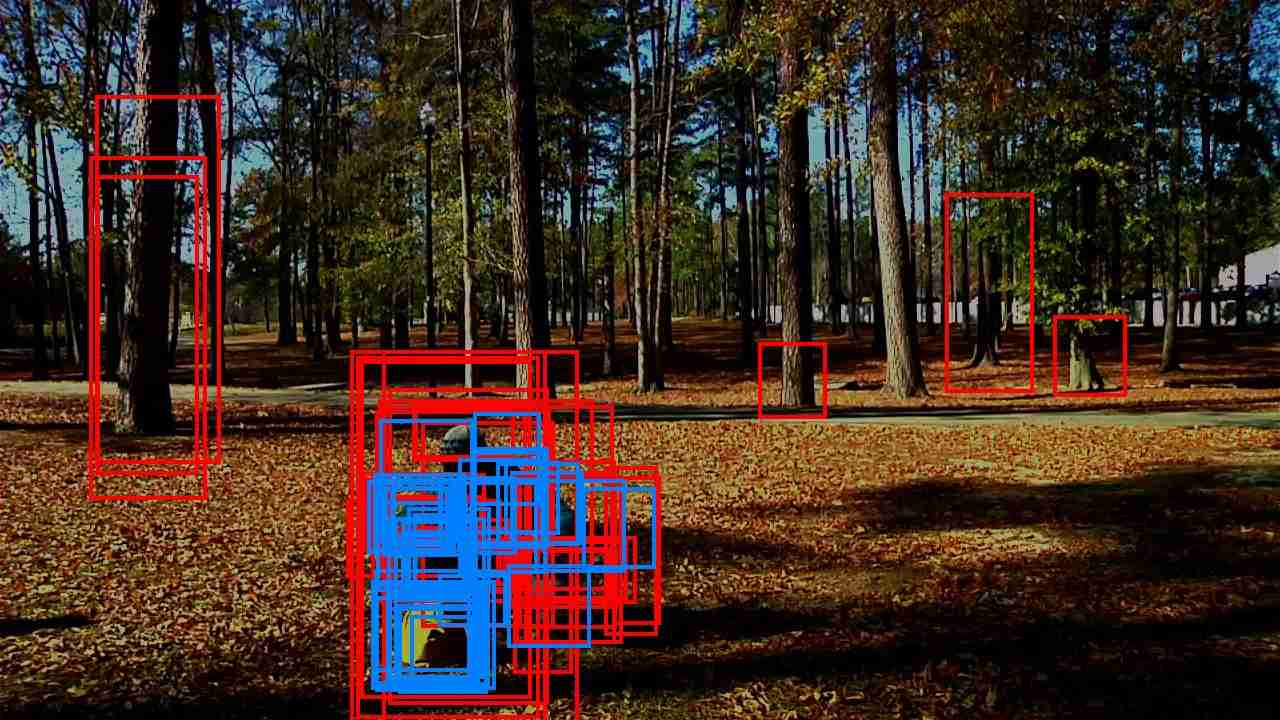}&
      \includegraphics[width=0.3\textwidth]{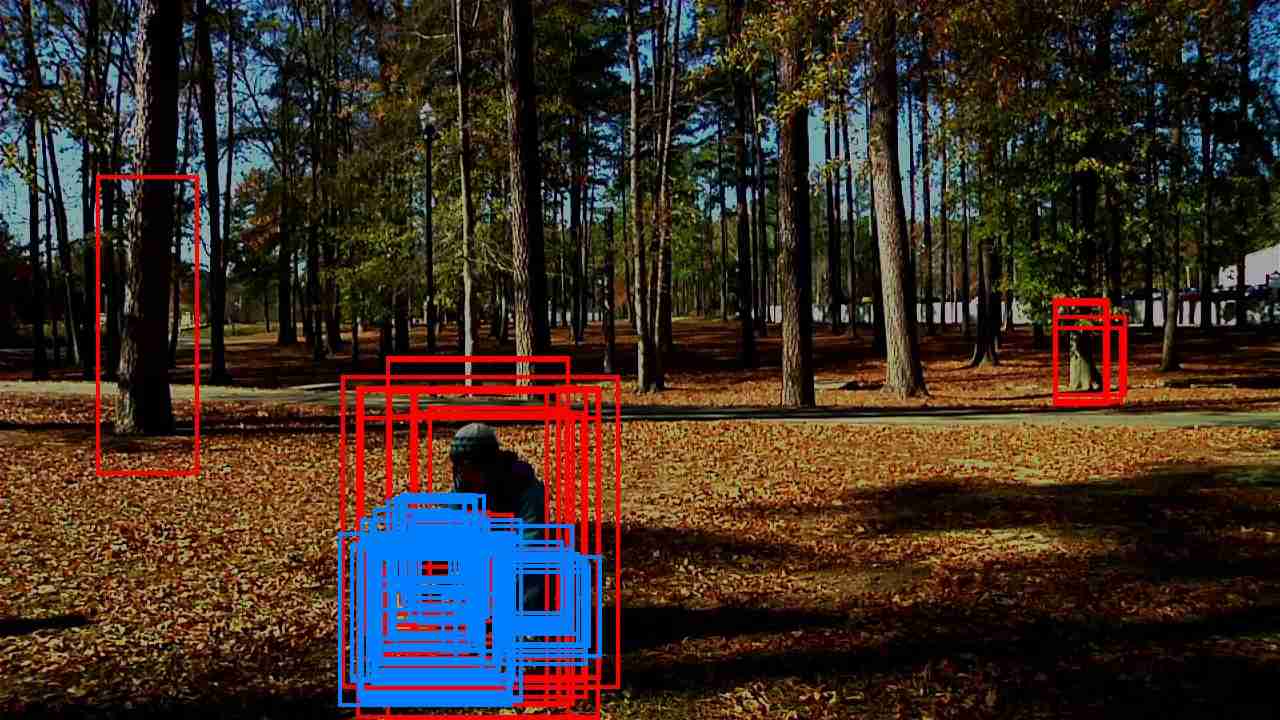}&
      \includegraphics[width=0.3\textwidth]{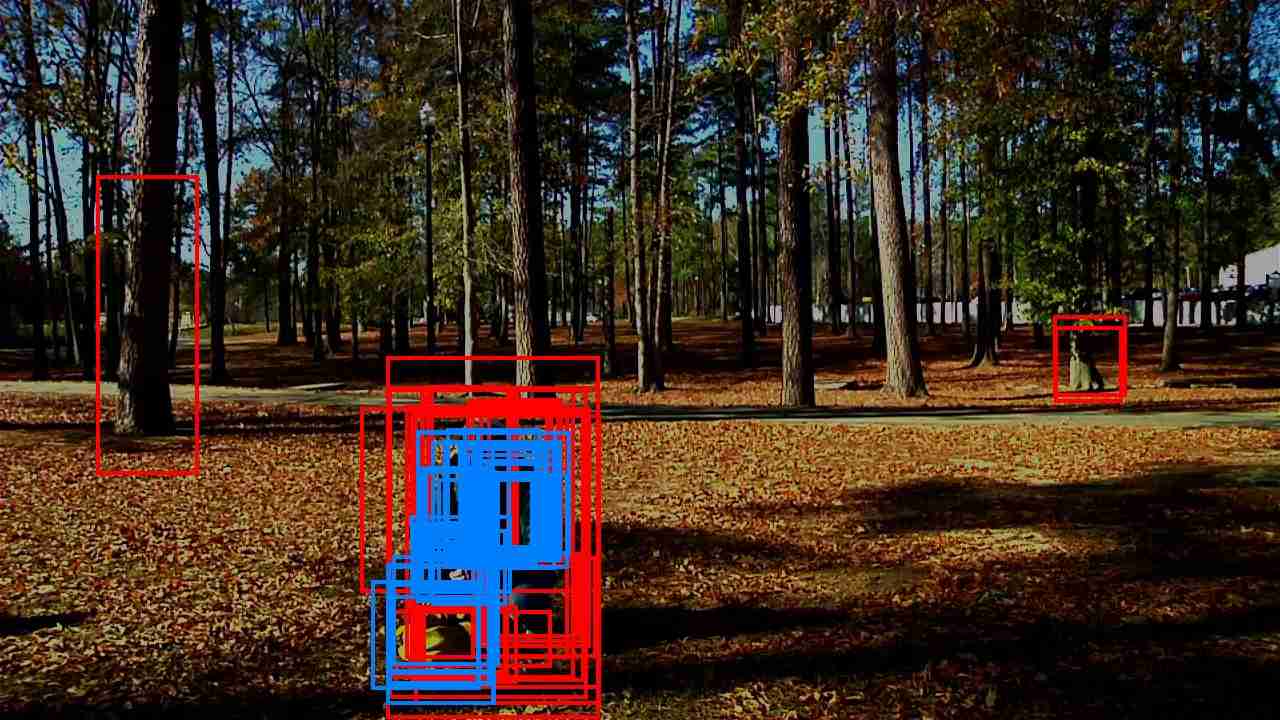}\\
      \raisebox{15pt}{(d)}&
      \includegraphics[width=0.3\textwidth]{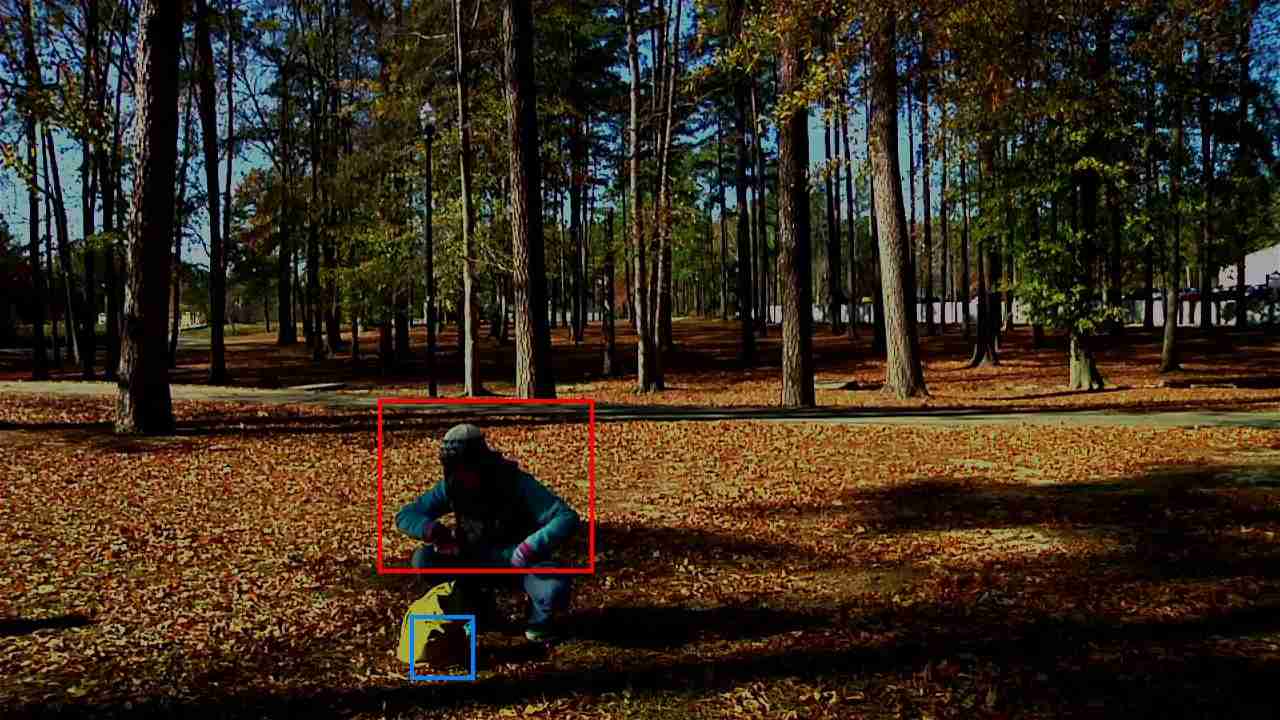}&
      \includegraphics[width=0.3\textwidth]{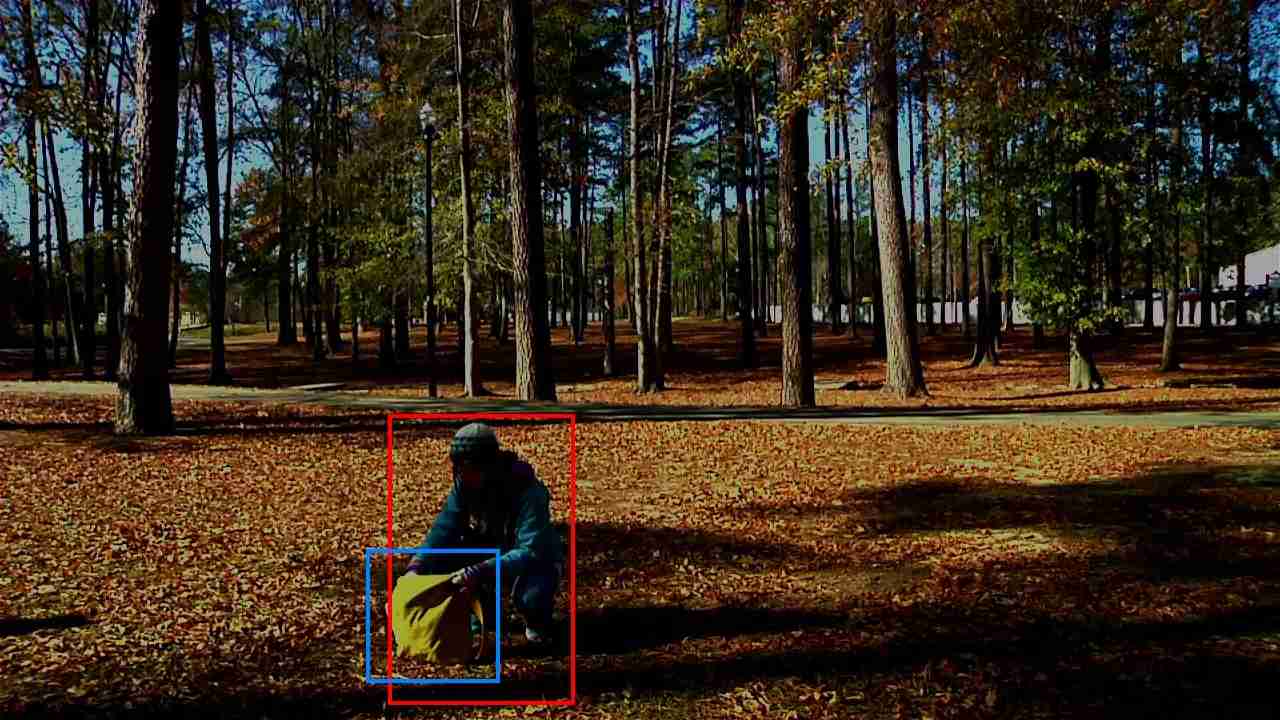}&
      \includegraphics[width=0.3\textwidth]{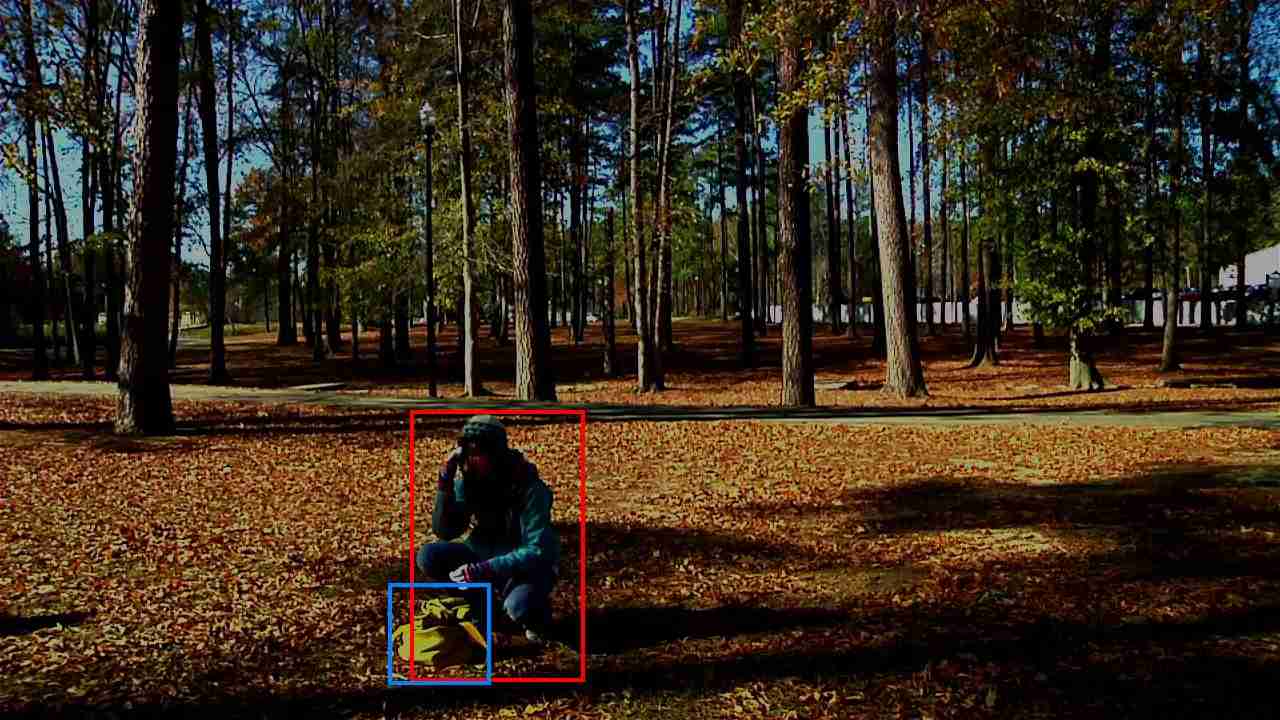}
    \end{tabular}
  \end{center}
  \caption{The operation of a detection-based tracker.
    (a)~Output of the detection sources, biased to yield false positives.
    (b)~The top-scoring output of the detection source.
    (c)~Augmenting the output of the detection sources with forward-projected
    detections.
    (d)~The optimal tracks selected by the Viterbi algorithm.}
  \label{fig:tracker}
\end{figure*}

The operation of a detection-based tracker is illustrated in
Fig.~\ref{fig:tracker}.
This example demonstrates several things of note.
First, reliable tracks are produced despite an unreliable detection source.
Second, the optimal track contains detections with suboptimal score.
Row~(b) demonstrates that selecting the top-scoring detection does not yield a
temporally-coherent track.
Third, forward-projection of detections from the second to third column in
row~(c) compensates for the lack of raw detections in the third column of
row~(a).

Detection-based tracking runs in time $O(TJ^2)$ on videos of length~$T$
with~$J$ detections per frame.
In practice, the run time is dominated by the detection process and the
dynamic-programming step.
Limiting~$J$ to a small number speeds up the tracker considerably while
minimally impacting track quality.
We further improve the speed of the detectors when running many object classes
by factoring the computation of the HOG pyramid.

\section{Evaluation of detection-based tracking}

We evaluated detection-base tracking using the year-one (Y1) corpus produced
by DARPA for the Mind's Eye program.
These videos are provided at 720p@30fps and range from 42 to 1727 frames in
length, with an average of 438.84 frames, and depict people interacting with a
variety of objects to enact common English verbs.

Four Mind's Eye teams (University at Buffalo, \citealt{buffalo2011}, Stanford
Research Institute, \citealt{sri2011}, University of California at Berkeley,
\citealt{berkeley2011}, and University of Southern California,
\citealt{usc2011}) independently produced human-annotated tracks for different
portions of Y1.
We used these sources of human-annotated tracks to evaluate the performance of
detection-based tracking by computing human-human intercoder agreement between
all pairs of the four sources of human-annotated tracks and human-machine
intercoder agreement between a detection-based tracker and all four of these
sources.
Since each team annotated different portions of Y1, each such intercoder
agreement measure was computed only over the~$N$ videos shared by each pair, as
reported in Table~\ref{tab:intercoder-agreement}(a).
One team (University at Buffalo, \citealt{buffalo2011}) annotated detections as
clusters of quadrilaterals around object parts.
These were converted to a single bounding box.

Different teams labeled the tracks with different class labels.
It was possible to determine from these labels whether the track was for a
person or nonperson by assuming that the labels `person' and `human', and only
those labels, denoted person tracks, but it was not possible to automatically
make finer-grained class comparisons.
Thus we independently compared person tracks with person tracks and nonperson
tracks with nonperson tracks.
When comparing an annotation~$u$ of a video~$n$ containing~$l^u_n$ person tracks
with an annotation~$v$ of that same video containing~$l^v_n$ person tracks, we
compared $\lfloor l^u_n,l^v_n\rfloor$ person tracks.
We selected the best over all $\lfloor l^u_n,l^v_n\rfloor!$ permutation
mappings~$\rho_n$ between person tracks in~$u$ and person tracks in~$v$.
A permutation mapping was preferred when it had higher average overlap score
among corresponding boxes across the tracks and the frames in a video, where
the overlap score was that used by the PASCAL VOC Challenge
\citep{Everingham10}, namely the ratio of the area of their intersection to the
area of their union.
Different tracks could annotate different frames of a video.
When comparing such, we considered only the shared frames.

For every pair of teams, we computed the mean and standard deviation of the
overlap score across all shared frames in all tracks in the best permutation
mappings for all shared videos.
The averaging process used both to determine the best permutation mapping for
each video pair and to determine overall mean and standard deviation measures
weighted each overlap score equally.
More precisely, if $1\leq n\leq N$, $1\leq l\leq\lfloor l^u_n,l^v_n\rfloor$
denotes a shared track for video~$n$, $T^l_n$ denotes the set of shared frames
for that shared track~$l$ in video~$n$, $U^n_t$ and $V^n_t$ denote the vector
of boxes for frame~$t$ in video~$n$ for annotations~$u$ and~$v$ respectively,
and~$O$ denotes the overlap measure, we score a permutation mapping~$\rho_n$
for video~$n$ as:
\begin{equation*}
  \cfrac{1}{\displaystyle\sum_{l=1}^{\lfloor l^u_n,l^v_n\rfloor}\lvert T^l_n\rvert}
  \sum_{l=1}^{\lfloor l^u_n,l^v_n\rfloor}\sum_{t\in T^l_n}
  O(\rho_n (U^n_t)[l],V^n_t[l])
\end{equation*}
and computed the mean overlap for a pair of teams as:
\begin{equation*}
  \cfrac{1}{\displaystyle
    \sum_{n=1}^N\sum_{l=1}^{\lfloor l^u_n,l^v_n\rfloor}\lvert T^l_n\rvert}
  \sum_{n=1}^N\sum_{l=1}^{\lfloor l^u_n,l^v_n\rfloor}\sum_{t\in T^l_n}
  O(\rho_n (U^n_t)[l],V^n_t[l])
\end{equation*}
with an analogous computation for standard deviation and nonperson tracks.

The overall mean and standard deviation measures, reported in
Table~\ref{tab:intercoder-agreement}(b,c), indicate that the mean human-human
overlap is only marginally greater than the mean human-machine overlap by about
one standard deviation.
This suggests that improvement in tracker performance is unlikely to lead to
significant improvement in action recognition performance and sentential
description quality.

\begin{table*}
    \begin{center}
      \scalebox{0.98}{
      \begin{tabular}{@{}ccc@{}}
	\begin{tabular}{l|rrrr}
	  $N$ & UB & SRI  & UCB  & USC \\\hline
	  UB  &    & 8    & 20   & 8   \\
	  SRI & 8  &      & 1201 & 95  \\
	  UCB & 20 & 1201 &      & 204 \\
	  USC & 8  & 95   & 204  &     \\\hline
	  us  & 48 & 1254 & 1829 & 360 \\
	\end{tabular}&
	\begin{tabular}{l|rrrr}
	  $\mu$ & UB   & SRI  & UCB  & USC  \\\hline
	  UB    &      & 0.76 & 0.68 & 0.59 \\
	  SRI   & 0.76 &      & 0.55 & 0.59 \\
	  UCB   & 0.68 & 0.55 &      & 0.48 \\
	  USC   & 0.59 & 0.59 & 0.48 &      \\\hline
	  us    & 0.54 & 0.40 & 0.35 & 0.43 \\
	\end{tabular}&
	\begin{tabular}{l|rrrr}
	  $\sigma$ & UB   & SRI  & UCB  & USC  \\\hline
	  UB       &      & 0.06 & 0.14 & 0.10 \\
	  SRI      & 0.06 &      & 0.27 & 0.16 \\
	  UCB      & 0.14 & 0.27 &      & 0.23 \\
	  USC      & 0.10 & 0.16 & 0.23 &      \\\hline
	  us       & 0.26 & 0.24 & 0.23 & 0.20 \\
	\end{tabular}\\
	(a) & (b) & (c)\\
      \end{tabular}}
    \end{center}
  \caption{(a)~The number of videos in common, (b)~the mean overlap, and (c)~the
    standard deviation in overlap between each pair of annotation sources. }
  \label{tab:intercoder-agreement}
\end{table*}

\section{Combining object detection and tracking}
\label{sec:detectionandtracking}

While detection-based tracking is resilient to low precision, it requires
perfect recall; it cannot generate a track through a frame that has no
detections and it cannot generate a track through a portion of the field of
view which has no detections regardless of how good the temporal-coherence of
the resulting track would be.
This brittleness means that any detection source employed will have to
significantly over-generate detections to achieve near-perfect recall.
This has a downside.
While the Viterbi algorithm has linear complexity in the number of frames, it
is quadratic in the number of detections per frame.
This drastically limits the number of detections that can reasonably be
processed leading to the necessity of tuning the thresholds on the detection
sources.
We have developed a novel mechanism to eliminate the need for a threshold and
track every possible detection, at every position and scale in the image, in
time linear in the number of detections and frames.
At the same time our approach eliminates the need for forward projection since
every detection is already present.
Our approach involves simultaneously performing object detection and tracking,
optimizing the joint object-detection and temporal-coherency score.

Our general approach is to compute the distance between pairs of detection
pyramids for adjacent frames, rather than using~$g$ to compute the distance
between pairs of individual detections.
These pyramids represent the set of all possible detections at all locations
and scales in the associated frame.
Employing a distance transform makes this process linear in the number of
location and scale positions in the pyramid.
Many detectors, \eg\ those of Felzenszwalb \etal, use such a scale-space
representation of frames to represent detections internally even though they
might not output such.
Our approach requires instrumenting such a detector to provide access to this
internal representation.

At a high-level, the Felzenszwalb \etal\ detectors learn a forest of HOG
\citep{freeman1995} filters for each object class along with their
characteristic displacements.
Detection proceeds by applying each HOG filter at every position in an image
pyramid followed by computing the optimal displacements at every position in
that image pyramid, thereby creating a new pyramid, the detection pyramid.
Finally, the detector searches the detection pyramid for high-scoring
detections and extracts those above a threshold.
The detector employs a dynamic-programming algorithm to efficiently compute
the optimal part displacements for the entire image pyramid.
This algorithm \citep{Felzenszwalb2010b} is very similar to the Viterbi
algorithm.
It is made tractable by the use of a generalized distance transform
\citep{Felzenszwalb2004} that allows it to scale linearly with the number of
image pyramid positions.
Given a set~$\mathcal{G}$ of points (which in our case denotes an image
pyramid), a distance metric~$d$ between pairs of points~$p$ and~$q$, and
an arbitrary function $\phi:\mathcal{G}\rightarrow\Re$, the generalized distance
transform~$D_{\phi}(q)$ computes:
\begin{equation*}
  D_{\phi}(q)=\min_{p\in\mathcal{G}}(d(p,q)+\phi(q))
\end{equation*}
in linear time for certain distance metrics including squared Euclidean
distance.

Instead of extracting and tracking just the thresholded detections, one can
directly track all detections in the entire pyramid simultaneously by defining
a distance measure between detection pyramids for adjacent frames and
performing the Viterbi tracking algorithm on these pyramids instead of sets of
detections in each frame.
To allow comparison between detections at different scales in the detection
pyramid, we convert the detection pyramid to a rectangular prism by
scaling the coordinates of the detections at scale~$s$ by~$\pi(s)$, chosen to
map the detection coordinates back to the coordinate system of the input frame.
We define the distance between two detections, $b$ and $b'$, in two detection
pyramids as a scaled squared Euclidean distance:
\begin{equation}
  d(b_{xys},b'_{x'y's'})=
  \begin{array}[t]{ll}
    (\pi(s)x-\pi(s')x')^2\\
    {}+(\pi(s)y-\pi(s')y')^2\\
    {}+\alpha(s-s')^2
  \end{array}
  \label{eq:distance}
\end{equation}
where~$x$ and~$y$ denote the original image coordinates of a detection center
at scale~$s$.
Nominally, detections are boxes.
Comparing two such boxes involves a four-dimensional distance metric.
However, with a detection pyramid, the aspect ratio of detections is fixed,
reducing this to a three-dimensional distance metric.
The coefficient~$\alpha$ in the distance metric weights a difference in
detection area differently than detection position.

The above amounts to replacing detections~$b^t_j$ with~$b^t_{xys}$, lattice
values~$\delta^t_j$ with~$\delta^t_{xys}$, and Eq.~\ref{eq:viterbi} with:
\begin{equation}
  \begin{array}[t]{@{}l@{\hspace*{-10pt}}}
    \textbf{for}\;x=1\;\textbf{to}\;X\\
    \textbf{do}\;\begin{array}[t]{@{}l@{}}
		 \textbf{for}\;y=1\;\textbf{to}\;Y\\
		 \textbf{do}\;\begin{array}[t]{@{}l@{}}
			      \textbf{for}\;s=1\;\textbf{to}\;S\;
			      \textbf{do}\;\delta^1_{xys}\assign
			      f(b^1_{xys})
			      \end{array}
		 \end{array}\\
    \textbf{for}\;t=2\;\textbf{to}\;T\\
    \textbf{do}\;\begin{array}[t]{@{}l@{}}
		 \textbf{for}\;x=1\;\textbf{to}\;X\\
		 \textbf{do}\;\begin{array}[t]{@{}l@{}}
			      \textbf{for}\;y=1\;\textbf{to}\;Y\\
 			      \textbf{do}\;\begin{array}[t]{@{}l@{}}
					   \textbf{for}\;s=1\;\textbf{to}\;S\\
					   \textbf{do}\;
					   \delta^t_{xys}\assign
					   \begin{array}[t]{@{}l@{}}
					   f(b^t_{xys})\\
					   +\displaystyle
					   \max_{x',y',s'}
					   g(b^{t-1}_{x'y's'},b^t_{xys})
					   +\delta^{t-1}_{x'y's'}
					   \end{array}
					   \end{array}
			      \end{array}
		 \end{array}
  \end{array}
  \label{eq:felzenszwalb-viterbi}
\end{equation}

The above formulation allows us to employ the generalized distance transform as
an analog to~$g$ in Eq.~\ref{eq:track}, although it restricts consideration
of~$g$ to be squared Euclidean distance rather than Euclidean distance.
We avail ourselves of the fact that the generalized distance transform operates
independently on each of the three dimensions~$x$, $y$, and~$s$ in order to
incorporate~$\alpha$ into Eq.~\ref{eq:distance}.
While linear-time use of the distance transform restricts the form of~$g$, it
places no restrictions on the form of~$f$.

One way to view the above is that the vector of~$\delta^t_j$ for all
$1\leq j\leq J_t$ from Eq.~\ref{eq:viterbi} is being represented as a pyramid
and the loop:
\begin{equation}
  \begin{array}[t]{@{}l@{}}
    \begin{array}[t]{@{}l@{}}
      \textbf{for}\;j=1\;\textbf{to}\;J_t\\
      \textbf{do}\;\delta^t_j\assign
      f(b^t_j)
      +\displaystyle
      \max_{j'=1}^{J_{t-1}} g(b^{t-1}_{j'},b^t_j)
      +\delta^{t-1}_{j'}
    \end{array}
  \end{array}
\end{equation}
is being performed as a linear-time construction of a generalized distance
transform rather than a quadratic-time nested pair of loops.
Another way to view the above is that we generalize the notion of a detection
pyramid from representing per-frame detections~$b_{xys}$ at three-dimensional
pyramid positions $\langle x,y,s\rangle$ to representing per-video
detections~$b^t_{xys}$ at four-dimensional pyramid positions
$\langle x,y,s,t\rangle$ and finding a sequence of per-video detections for
$1\leq~t\leq~T$ that optimizes the following variant of Eq.~\ref{eq:track}:
\begin{equation}
  \max_{\substack{x_1,\ldots,x_T\\y_1,\ldots,y_T\\s_1,\ldots,s_T}}
  \sum_{t=1}^T f(b^t_{x_ty_ts_t})+
  \sum_{t=2}^T g(b^{t-1}_{x_{t-1}y_{t-1}s_{t-1}},b^t_{x_ty_ts_t})
  \label{eq:fv}
\end{equation}
This combination of the detector and the tracker is performing simultaneous
detection and tracking integrating the information between the two.
Before, the tracker was affected by the detector but the detector was
unaffected by the tracker: potential low-scoring but temporally-coherent
detections would not even be generated by the detector despite the fact that
they would yield good tracks.
Because now, the detector no longer chooses which detections to produce but
instead scores all detections at every position and scale, the tracker is
able to choose among any possible detection.
Such tight integration of higher- and lower-level information will be revisited
when integrating event models into this framework.

\section{Combining tracking and event detection}
\label{sec:trackingandevents}

It is popular to use Hidden Markov Models (HMMs) to perform event
recognition \citep{Siskind1996, Starner98, Wang2009, Xu2002, Xu2005}.
When doing so, the log likelihood of a video conditioned on an event model is:
\begin{equation*}
  \log\sum_{k_1,\ldots,k_T}
  \exp\sum_{t=1}^Th(k_t,b^t_{j^{*}_t})+\sum_{t=2}^Ta(k_t,k_{t-1})
\end{equation*}
where~$k_t$ denotes the state of the HMM for frame~$t$, $h(k,b)$ denotes
the log probability of generating a detection~$b$ conditioned on being in
state~$k$, $a(k,k')$ denotes the log probability of transitioning from state~$k$
to~$k'$, and $j^{*}_t$ denotes index of the detection produced by the tracker
in frame~$t$.
This log likelihood can be computed with the forward algorithm \citep{Baum1966}
which is analogous to the Viterbi algorithm.
Maximum likelihood (ML), the standard approach to using HMMs for classification,
selects the event model that maximizes the likelihood of an observed event.
One can instead select the model with the maximum \emph{a posteriori}
(log) probability (MAP).
\begin{equation}
  \max_{k_1,\ldots,k_T}
  \sum_{t=1}^Th(k_t,b^t_{j^{*}_t})+
  \sum_{t=2}^Ta(k_t,k_{t-1})
  \label{eq:map}
\end{equation}
This can be computed with the Viterbi algorithm.
The advantage of doing so is that one can combine the Viterbi algorithm used for
detection-based tracking with the Viterbi algorithm used for event
classification.

One can combine Eq.~\ref{eq:track} with Eq.~\ref{eq:map} to yield a unified
cost function:
\begin{equation}
  \max_{j_1,\ldots,j_T}
  \max_{k_1,\ldots,k_T}
  \begin{array}[t]{l}
  \displaystyle\sum_{t=1}^T f(b^t_{j_t})
  +\displaystyle\sum_{t=2}^T g(b^{t-1}_{j_{t-1}},b^t_{j_t})\\
  +\displaystyle\sum_{t=1}^Th(k_t,b^t_{j_t})
  +\displaystyle\sum_{t=2}^Ta(k_t,k_{t-1})
  \end{array}
  \label{eq:unified}
\end{equation}
that computes the joint MAP of the best possible track and the best possible
state sequence by replacing $j^{*}_t$ with~$j_t$ inside nested quantification.
This too can be computed with the Viterbi algorithm, taking the lattice
values~$\delta^t_{jk}$ to be indexed by the detection index~$j$ and the
state~$k$, forming the cross product of the tracker lattice nodes and the
event lattice nodes:
\begin{equation}
  \begin{array}[t]{@{}l@{\hspace*{-20pt}}}
    \textbf{for}\;j=1\;\textbf{to}\;J_1\\
    \textbf{do}\;\begin{array}[t]{@{}l@{}}
		 \textbf{for}\;k=1\;\textbf{to}\;K\;
		 \textbf{do}\;\delta^1_{jk}\assign f(b^1_j)+h(k,b^1_j)
		 \end{array}\\
    \textbf{for}\;t=2\;\textbf{to}\;T\\
    \textbf{do}\;\begin{array}[t]{@{}l@{}}
		 \textbf{for}\;j=1\;\textbf{to}\;J_t\\
		 \textbf{do}\;\begin{array}[t]{@{}l@{}}
			      \textbf{for}\;k=1\;\textbf{to}\;K\\
			      \textbf{do}\;\begin{array}[t]{@{}l@{}}
					   \delta^t_{jk}\assign
					   \begin{array}[t]{l}
					   f(b^t_j)+h(k,b^t_j)\\
					   {}+\displaystyle
					   \max_{j'=1}^{J_{t-1}}
					   \max_{k'=1}^K
					   \begin{array}[t]{l}
					   g(b^{t-1}_{j'},b^t_j)
					   +a(k,k')\\
					   {}+\delta^{t-1}_{j'k'}
					   \end{array}
					   \end{array}
					   \end{array}
			      \end{array}
		 \end{array}
  \end{array}
  \label{eq:viterbisquared}
\end{equation}
This finds the optimal path through a graph where the nodes at every frame
represent the cross product of the detections and the HMM states.

Doing so performs simultaneous tracking and event classification.
Before, the event classifier was affected by the tracker but the tracker was
unaffected by the event classifier: potential low-scoring tracks would not
even be generated by the tracker despite the fact that they would yield a high
MAP estimate for some event class.
Because now, the tracker no longer chooses which tracks to produce but
instead scores all tracks, the event classifier is able to choose among any
possible track.
This amounts to a different kind of track-coherence measure that is tuned to
specific events.
Such a measure might otherwise be difficult to achieve without top-down
information from the event classifier.
For example applying this method to a video of a running person along with
an event model for running, will be more likely to compose a track out of
person detections that has high velocity and low change in direction.

Processing each frame~$t$ with the algorithm in Eq.~\ref{eq:viterbisquared} is
quadratic in~$J_tK$.
This can be problematic since~$J_tK$ can be large.
As before, we can make this linear in~$J_t$ using a generalized distance
transform.
One can make this linear in~$K$ for suitable state-transition functions~$a$
\citep{FelzenszwalbHK03}.

Two practical issues arise when applying the above method.
First, one can factor Eq.~\ref{eq:before} as Eq.~\ref{eq:after}:
\begin{equation}
  \displaystyle\max_{j'=1}^{J_{t-1}}
  \displaystyle\max_{k'=1}^K
  \left(g(b^{t-1}_{j'},b^t_j)+a(k,k')+\delta^{t-1}_{j'k'}\right)
  \label{eq:before}
\end{equation}
\begin{equation}
  \displaystyle\max_{j'=1}^{J_{t-1}}
  \left(g(b^{t-1}_{j'},b^t_j)+
  \displaystyle\max_{k'=1}^K
  \left(a(k,k')+\delta^{t-1}_{j'k'}\right)\right)
  \label{eq:after}
\end{equation}
\noindent
This is important because the computation of $g(b^{t-1}_{j'},b^t_j)$ might
be expensive as it involves a projection of $b^{t-1}_{j'}$ forward one frame
(\eg\ using optical flow or KLT).\@
Second, when applying this method to multiple event models, the same
factorization can be extended to cache the computation of
$g(b^{t-1}_{j'},b^t_j)$ across different event models as this term does not
depend on the event model.

\section{Combining object detection, tracking and event detection}
\label{sec:detectionandtrackingandevents}

One can combine the methods of Sections~\ref{sec:detectionandtracking}
and~\ref{sec:trackingandevents} to optimize a cost function:
\begin{equation}
  \hspace{-0.5ex}
  \max_{\substack{x_1,\ldots,x_T\\y_1,\ldots,y_T\\s_1,\ldots,s_T\\k_1,\ldots,k_T}}
  \begin{array}[t]{l@{\hspace*{-10pt}}}
  \displaystyle\sum_{t=1}^T f(b^t_{x_ty_ts_t})+h(k_t,b^t_{x_ty_ts_t})\\
  +\displaystyle\sum_{t=2}^T g(b^{t-1}_{x_{t-1}y_{t-1}s_{t-1}},b^t_{x_ty_ts_t})
  +a(k_t,k_{t-1})
  \end{array}
  \label{eq:allthree}
\end{equation}
that combines Eq.~\ref{eq:fv} with Eq.~\ref{eq:unified} by forming a large
Viterbi lattice with values~$\delta^t_{xysk}$.

One practical issue arises when applying the above method.
In Eq.~\ref{eq:allthree}, $h$~is a function of~$b^t_{x_ty_ts_t}$, the detection in
the current frame.
This allows the HMM event model to depend on static object characteristics such
as position, shape, and pose.
However, many approaches to event recognition using HMMs use temporal
derivatives of such characteristics to provide object velocity and acceleration
information \citep{Siskind1996, Starner98}.
Having~$h$ also be a function of~$b^{t-1}_{x_{t-1}y_{t-1}s_{t-1}}$, the
detection in the previous frame, requires incorporation~$h$ into the
generalized distance transform and thus restricts its form.

The above combination performs simultaneous object detection, tracking, and
event classification, integrating information across all three.
Without such information integration, the object detector is unaffected by the
tracker which is in turn unaffected by the event model.
With such integration, the event model can influence the tracker and both can
influence the object detector.

This is important because current object detectors cannot reliably detect
small, deformable, or partially occluded objects.
Moreover, current trackers also fail to track such objects.
Information from the event model can focus the object detector and tracker on
those particular objects that participate in a specified event.
An event model for recognizing an agent picking an object up can bias the
object detector and tracker to search for an object that exhibits a particular
profile of motion relative to the agent, namely where the object is in close
proximity to the agent, the object starts out being at rest while the agent
approaches the object, then the agent touches the object, followed by the
object moving with the agent.

A traditional view of the relationship between object and event detection
suggests that one recognizes a \emph{hammering} event, in part, because one
detects a \emph{hammer}.
Our unified approach inverts the traditional view, suggesting that one can
recognize a \emph{hammer}, in part, by detecting a \emph{hammering} event.
Furthermore, a strength of our approach is that such relationships are not
encoded explicitly, do not have to be annotated in the training data for the
event models, and are learned automatically as part of learning the parameters
of the different event models.
This is to say that the relationship between a person and the objects they
manipulate can be learned from the co-occurrence of tracks in the training
data, rather than from manually annotated symbolic relationships.

\vspace{-1ex}
\section{Experimental results}
\begin{figure*}[p]
  \begin{center}
    \begin{tabular}{@{}c@{\hspace*{2pt}}c@{\hspace*{2pt}}c@{}}
      \includegraphics[width=0.32\textwidth]{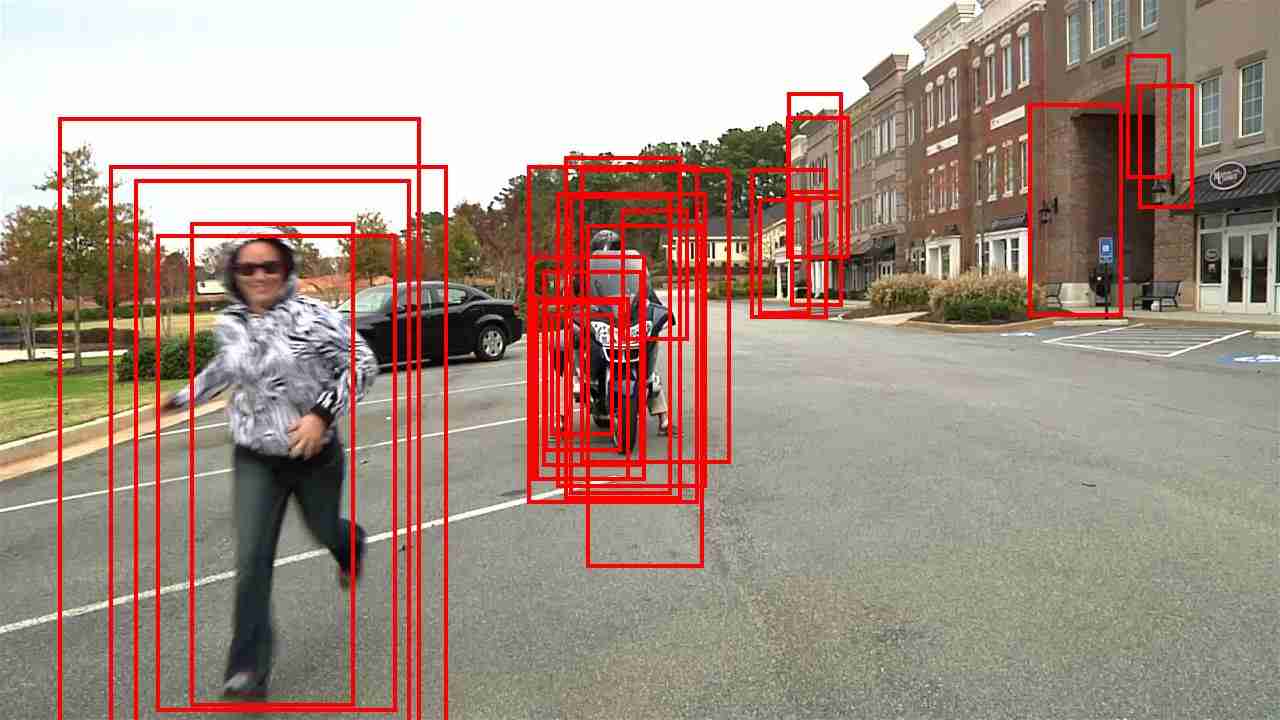}&
      \includegraphics[width=0.32\textwidth]{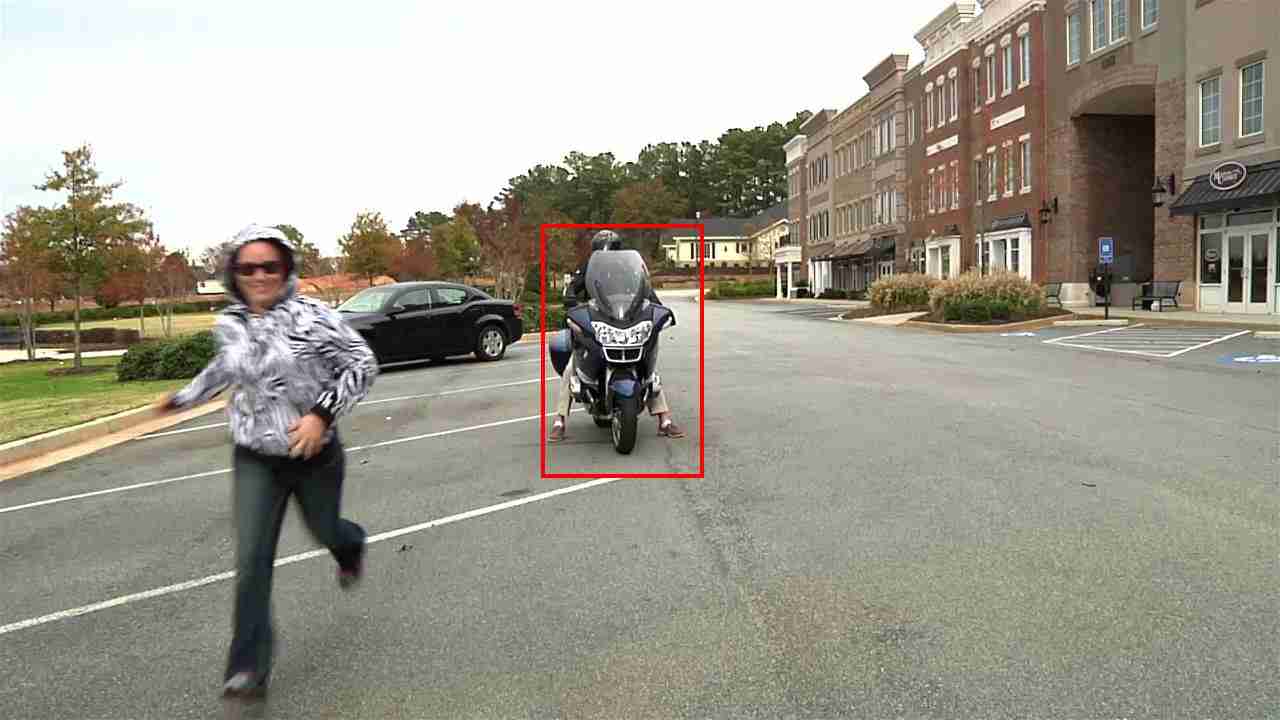}&
      \includegraphics[width=0.32\textwidth]{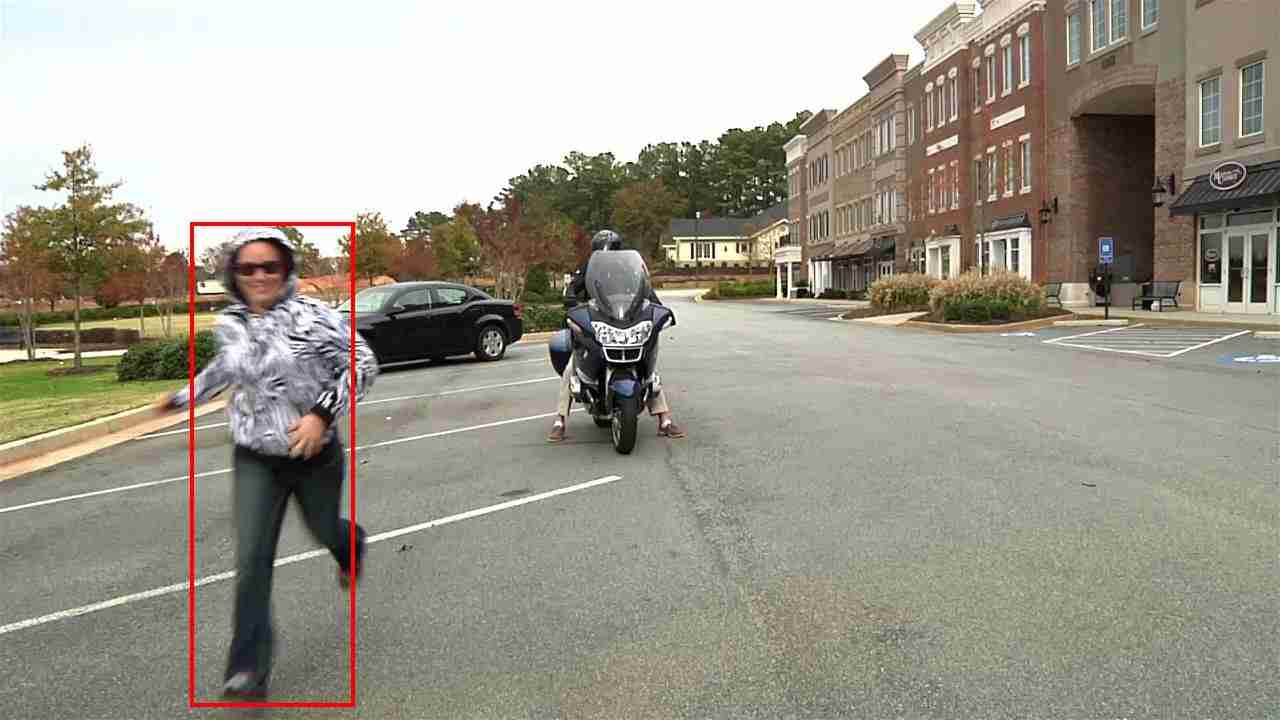}\\
      \includegraphics[width=0.32\textwidth]{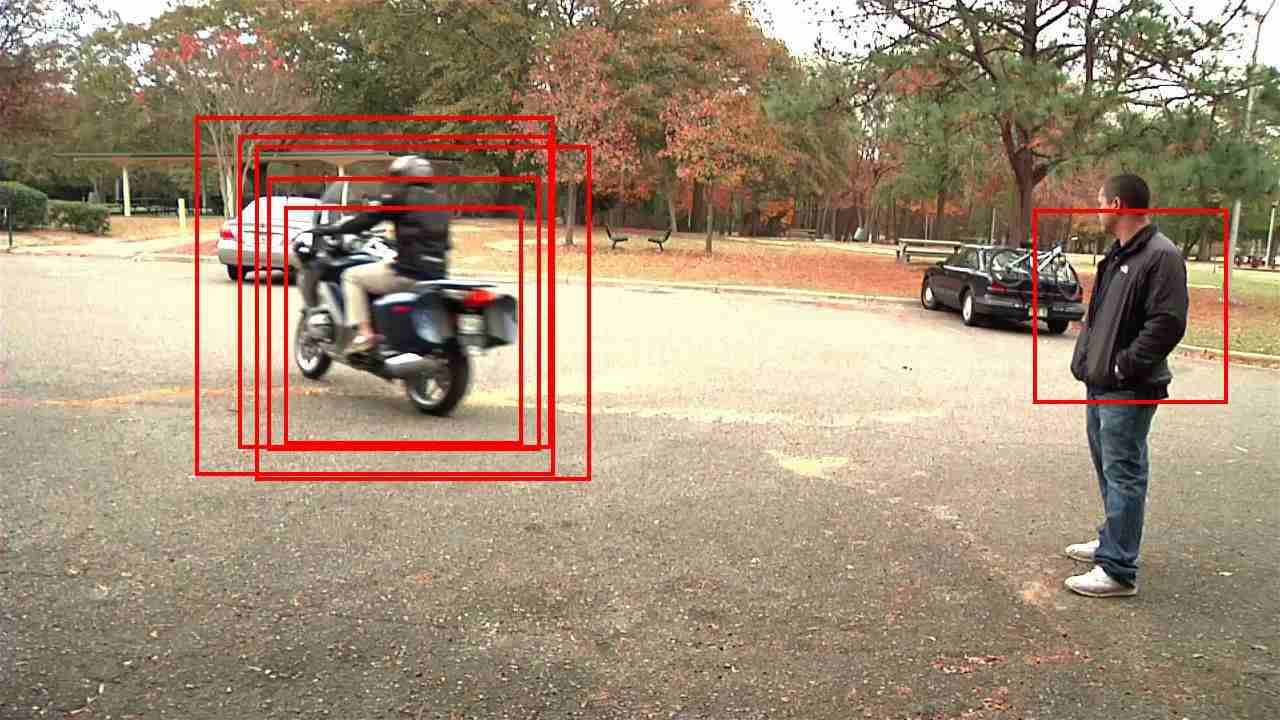}&
      \includegraphics[width=0.32\textwidth]{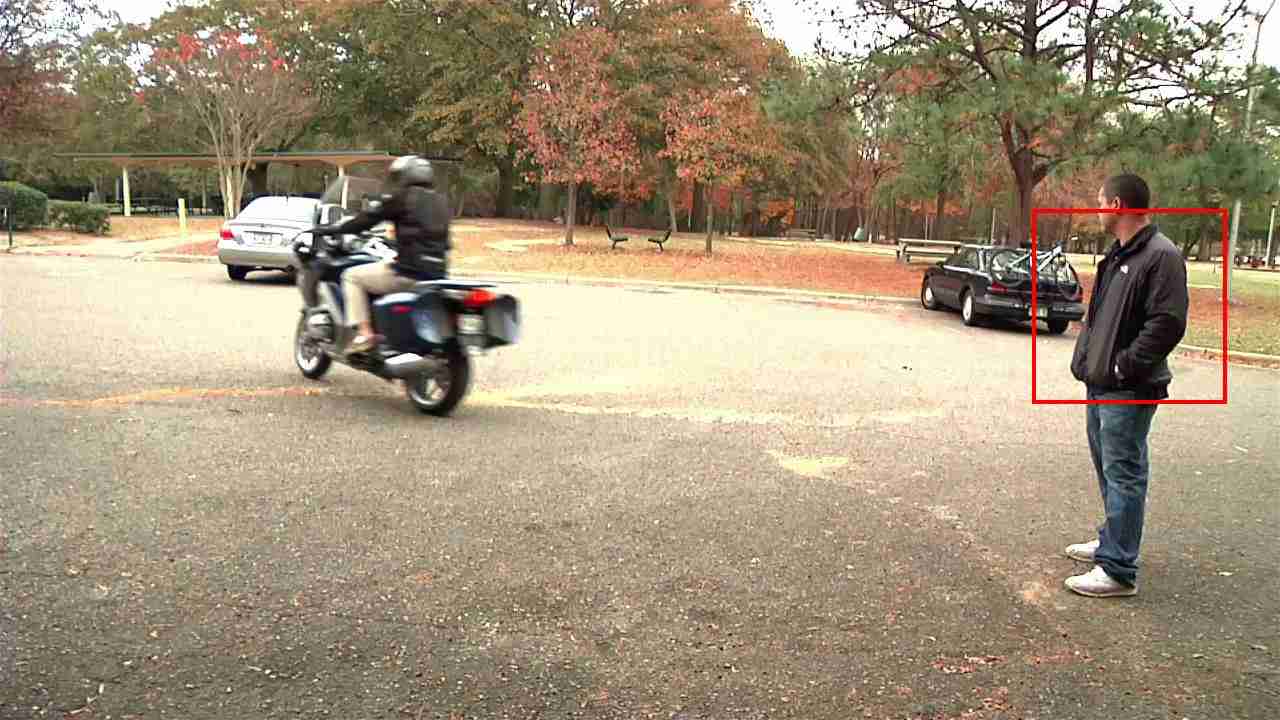}&
      \includegraphics[width=0.32\textwidth]{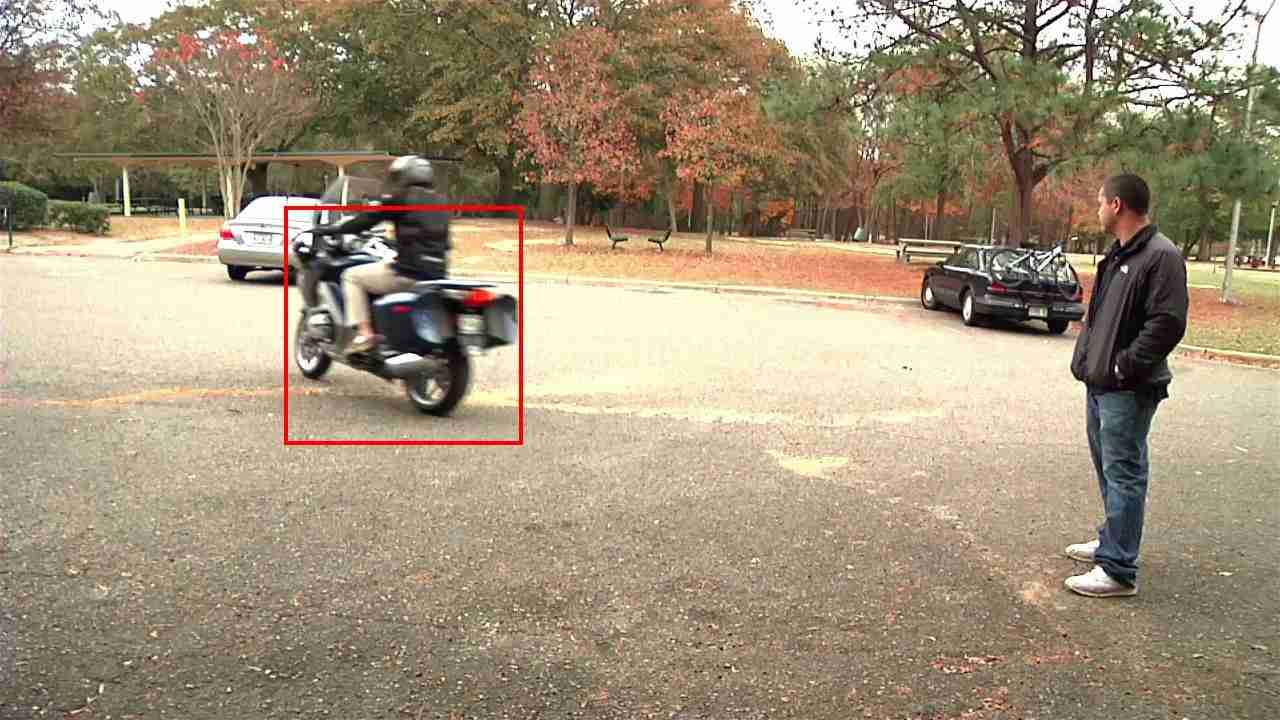}\\
      \includegraphics[width=0.32\textwidth]{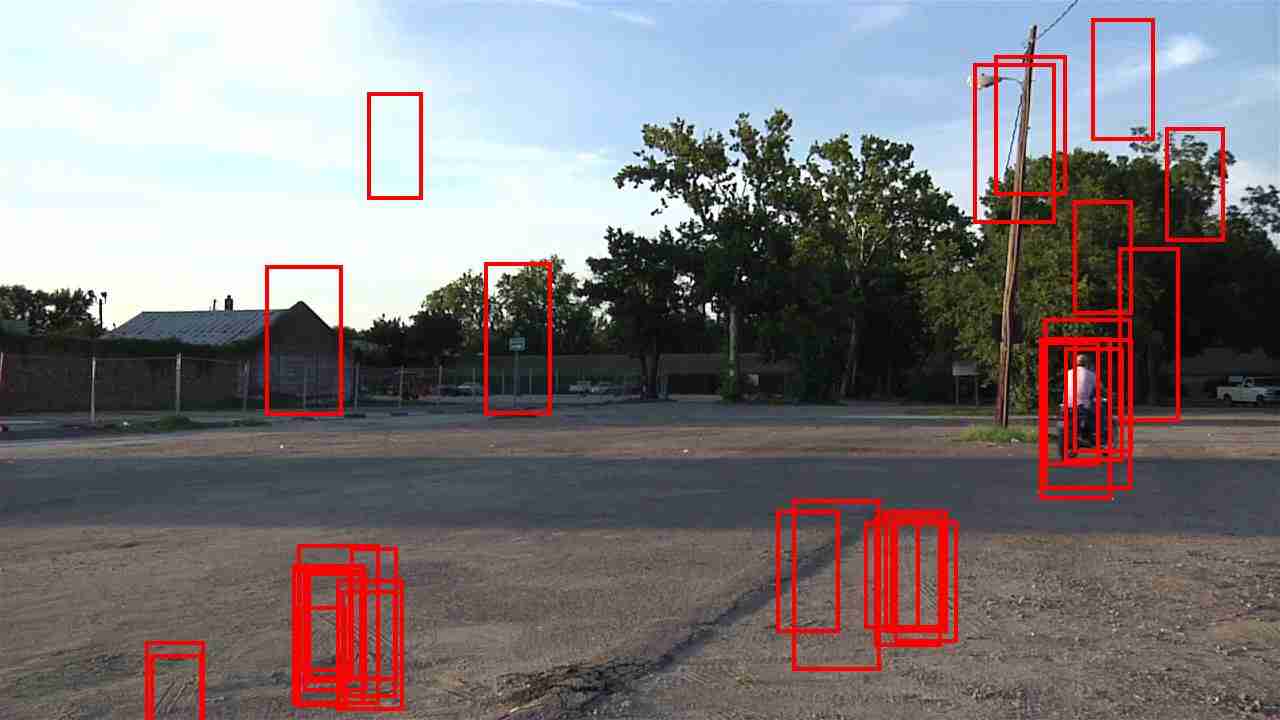}&
      \includegraphics[width=0.32\textwidth]{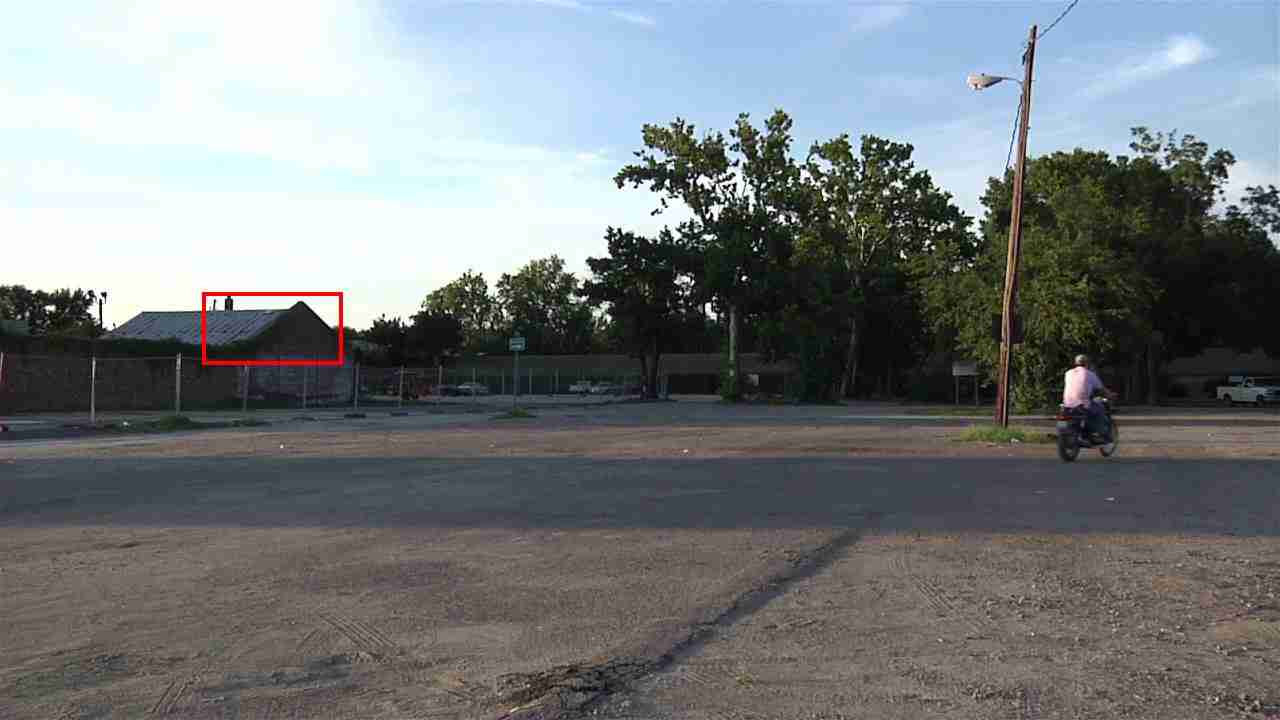}&
      \includegraphics[width=0.32\textwidth]{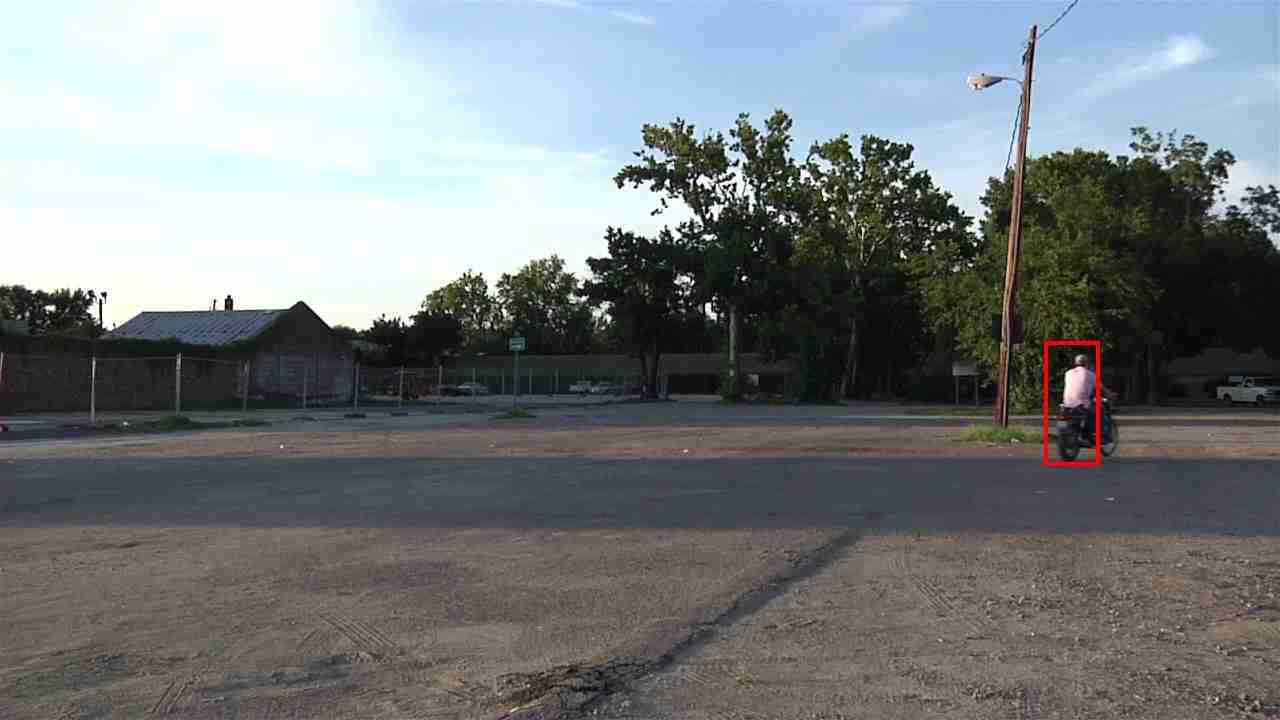}\\
      \includegraphics[width=0.32\textwidth]{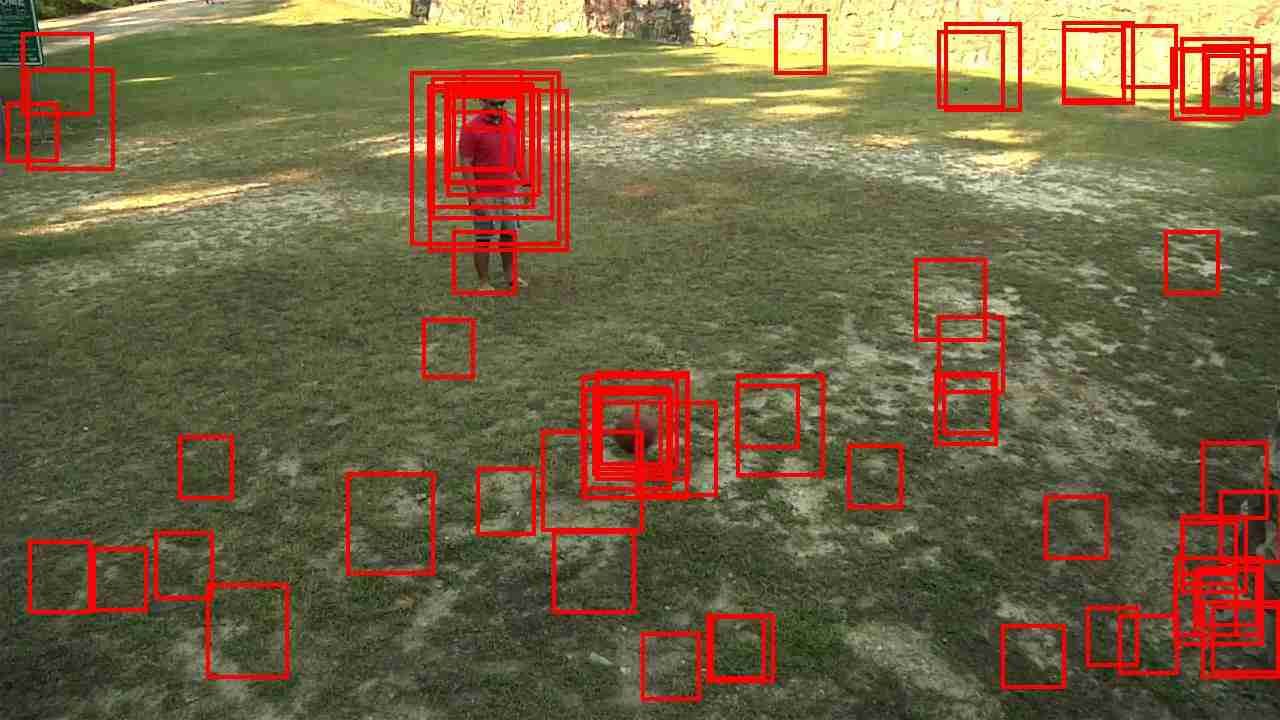}&
      \includegraphics[width=0.32\textwidth]{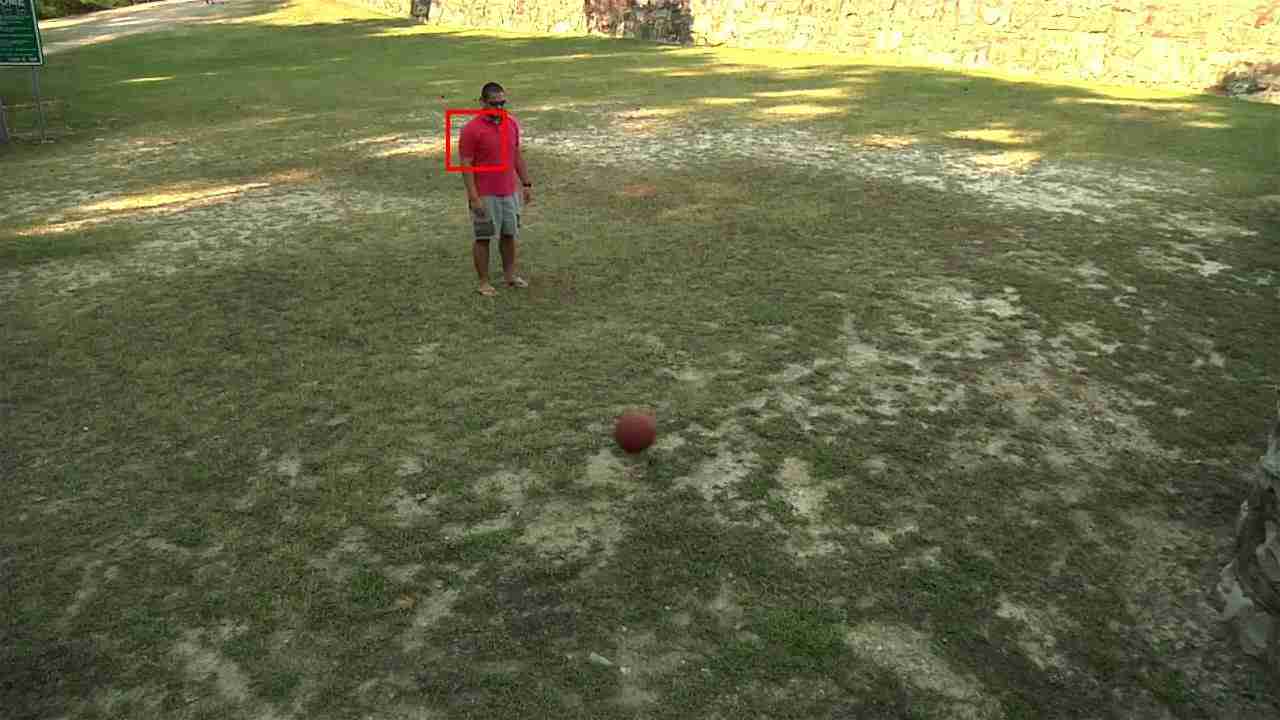}&
      \includegraphics[width=0.32\textwidth]{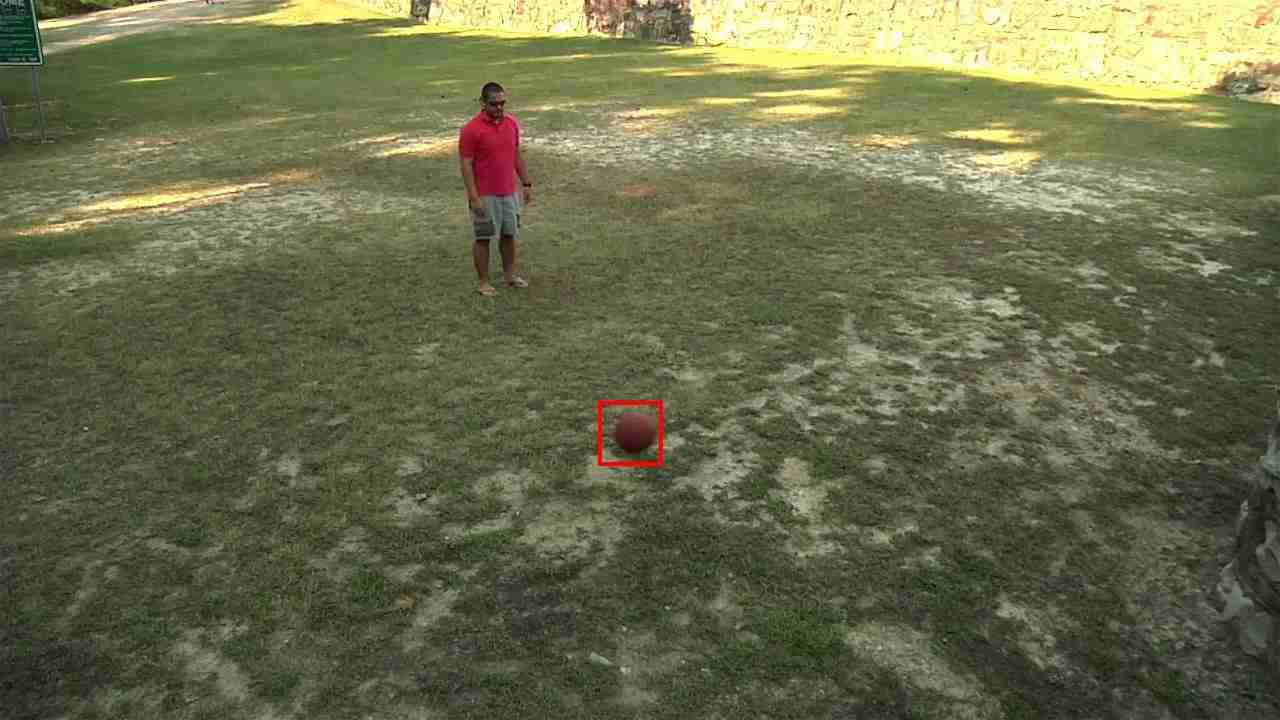}\\
      (a)&(b)&(c)\\
    \end{tabular}
  \end{center}
  \caption{Improved performance of simultaneous object detection and tracking.
    (a)~Output of the Felzenszwalb \etal\ detector.
    (b)~Tracks produced by detection-based tracking.
    (c)~Tracks produced by simultaneous object detection and tracking.}
  \label{fig:objectsandtracking}
\end{figure*}

\vspace{-1ex}
Figure \ref{fig:objectsandtracking} demonstrates improved performance of
simultaneous object detection and tracking~(c) over object detection~(a) and
tracking~(b) in isolation.
This happens for different reasons: motion blur, even for large objects, can
lead to poor detection results and hence poor tracks, small objects are
difficult to detect and track, and integration can improve detection and
tracking of deformable objects, such as a person transitioning from an upright
pose to sitting down.

\begin{figure*}[p]
  \begin{center}
    \begin{tabular}{@{}c@{\hspace*{2pt}}c@{\hspace*{2pt}}c@{}}
      \includegraphics[width=0.32\textwidth]{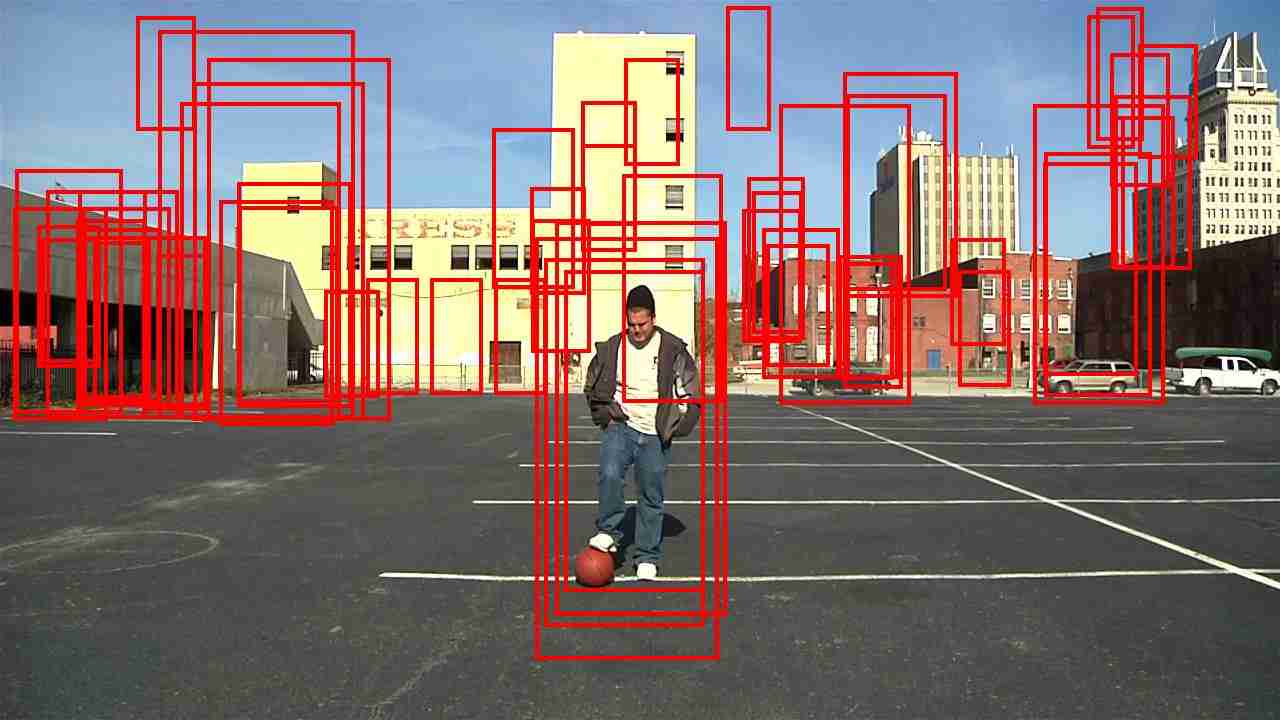}&
      \includegraphics[width=0.32\textwidth]{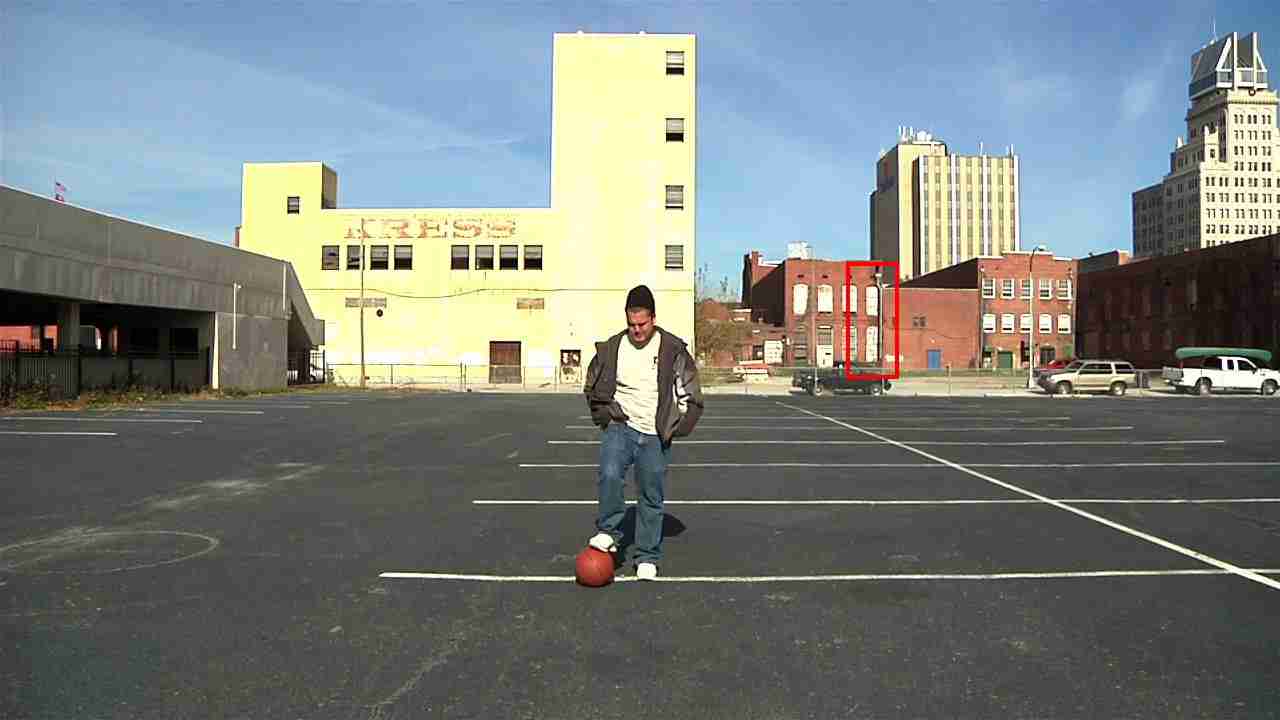}&
      \includegraphics[width=0.32\textwidth]{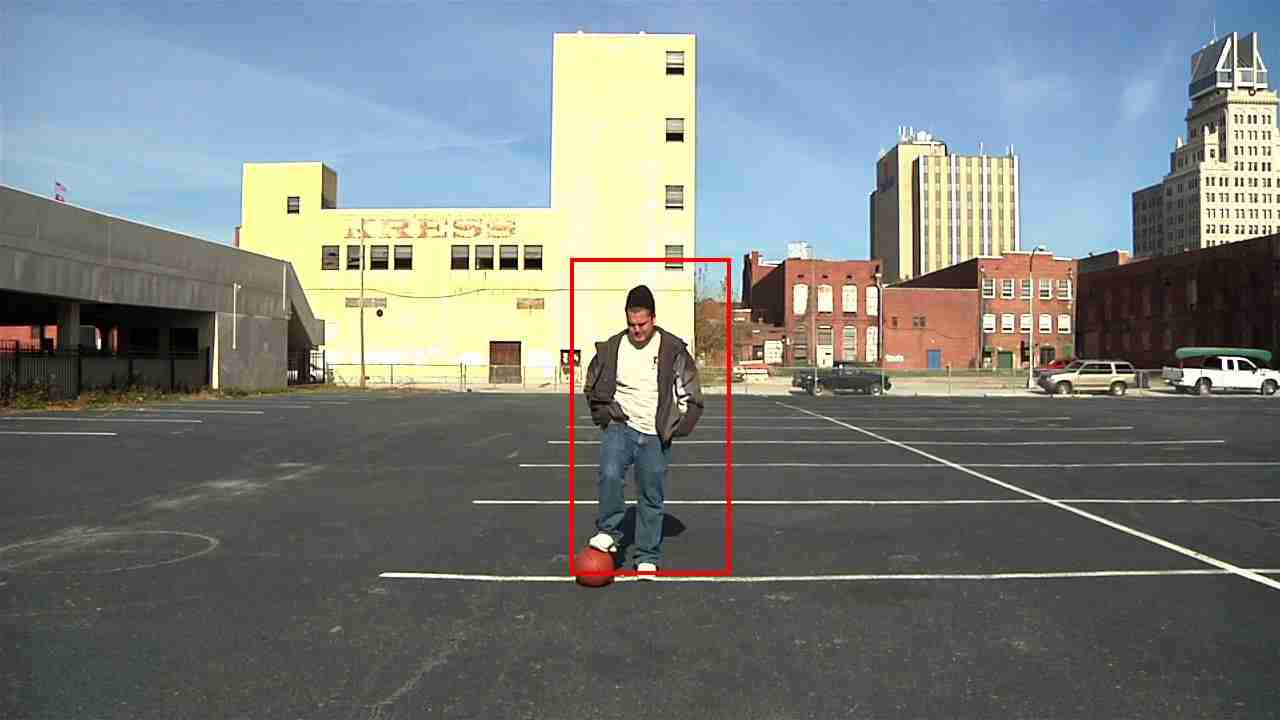}\\
      \includegraphics[width=0.32\textwidth]{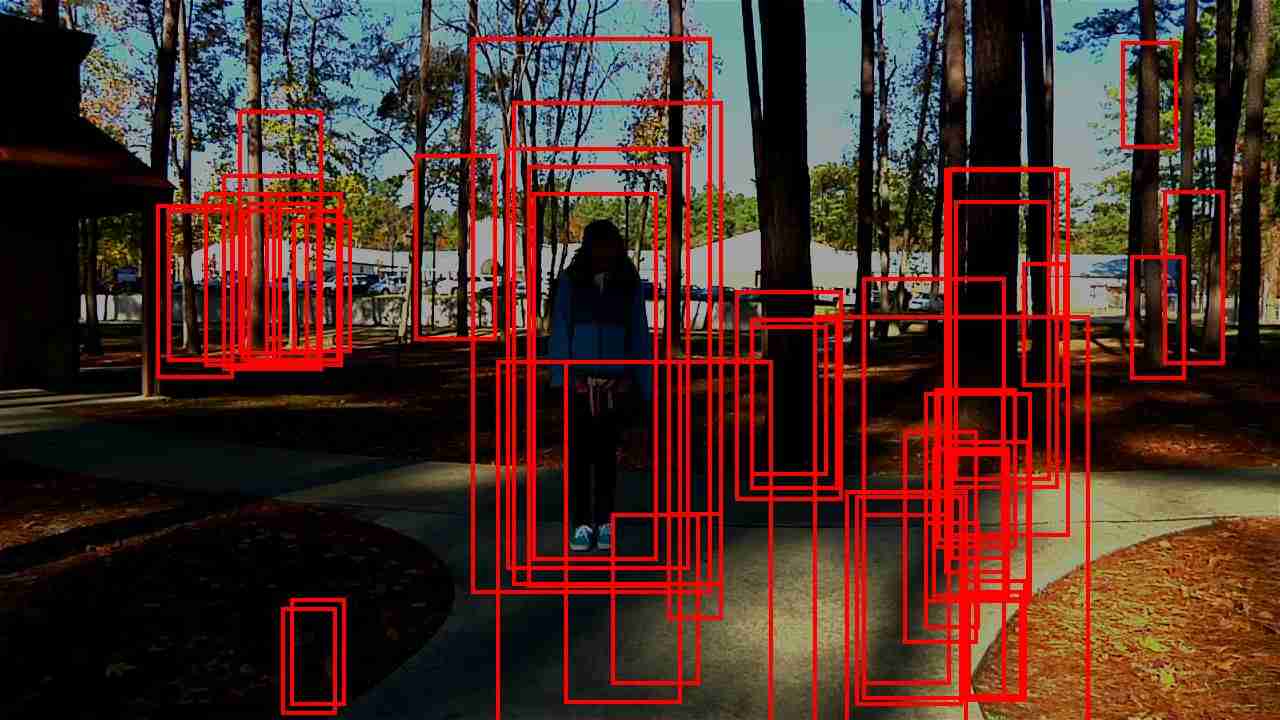}&
      \includegraphics[width=0.32\textwidth]{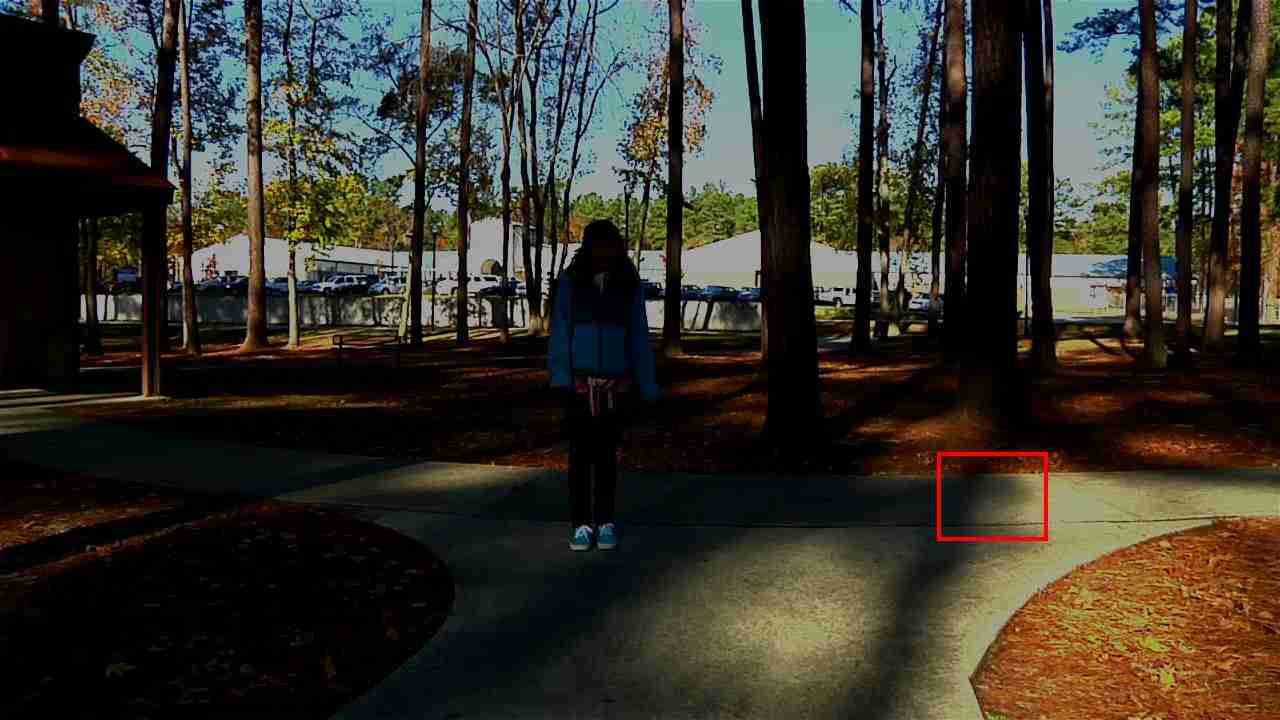}&
      \includegraphics[width=0.32\textwidth]{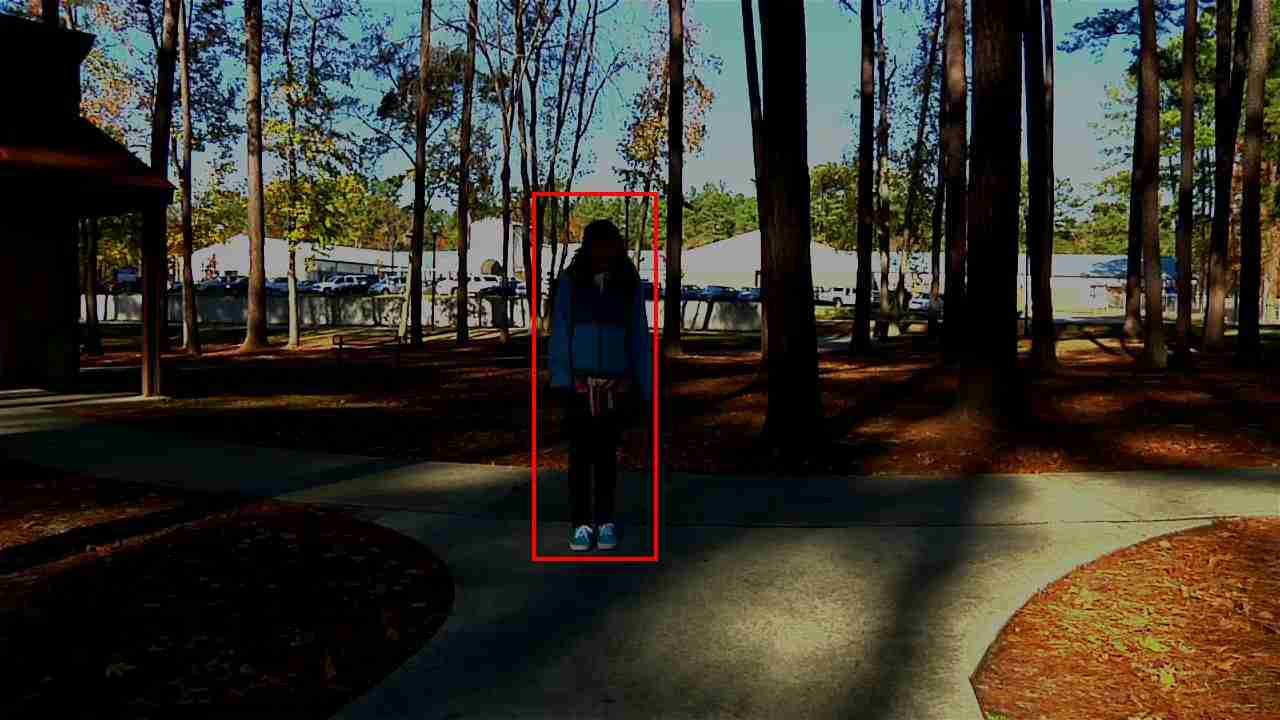}\\
      \includegraphics[width=0.32\textwidth]{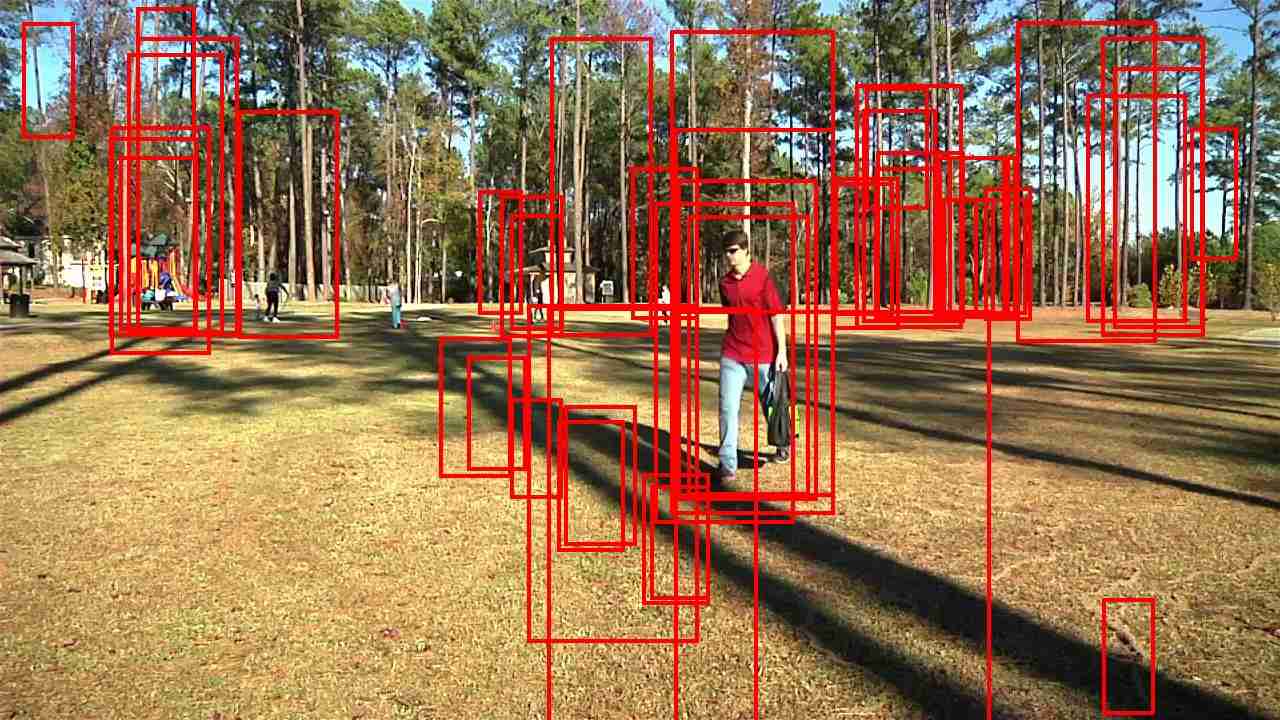}&
      \includegraphics[width=0.32\textwidth]{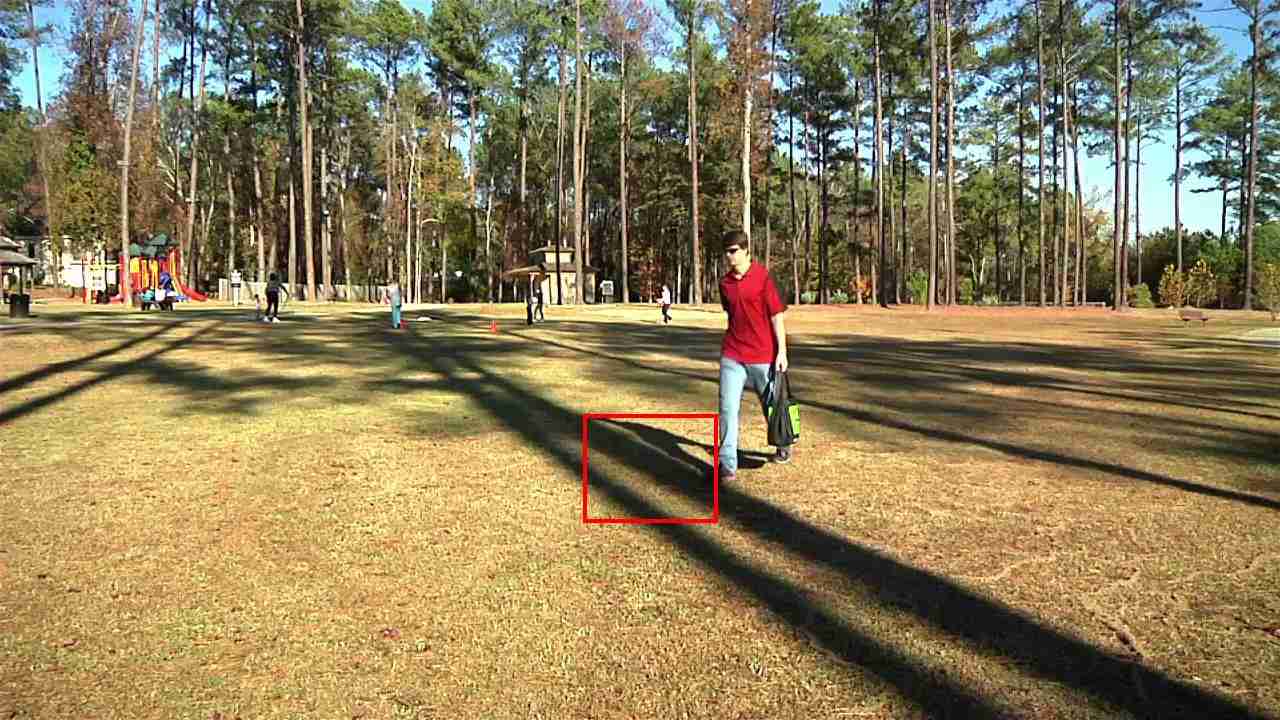}&
      \includegraphics[width=0.32\textwidth]{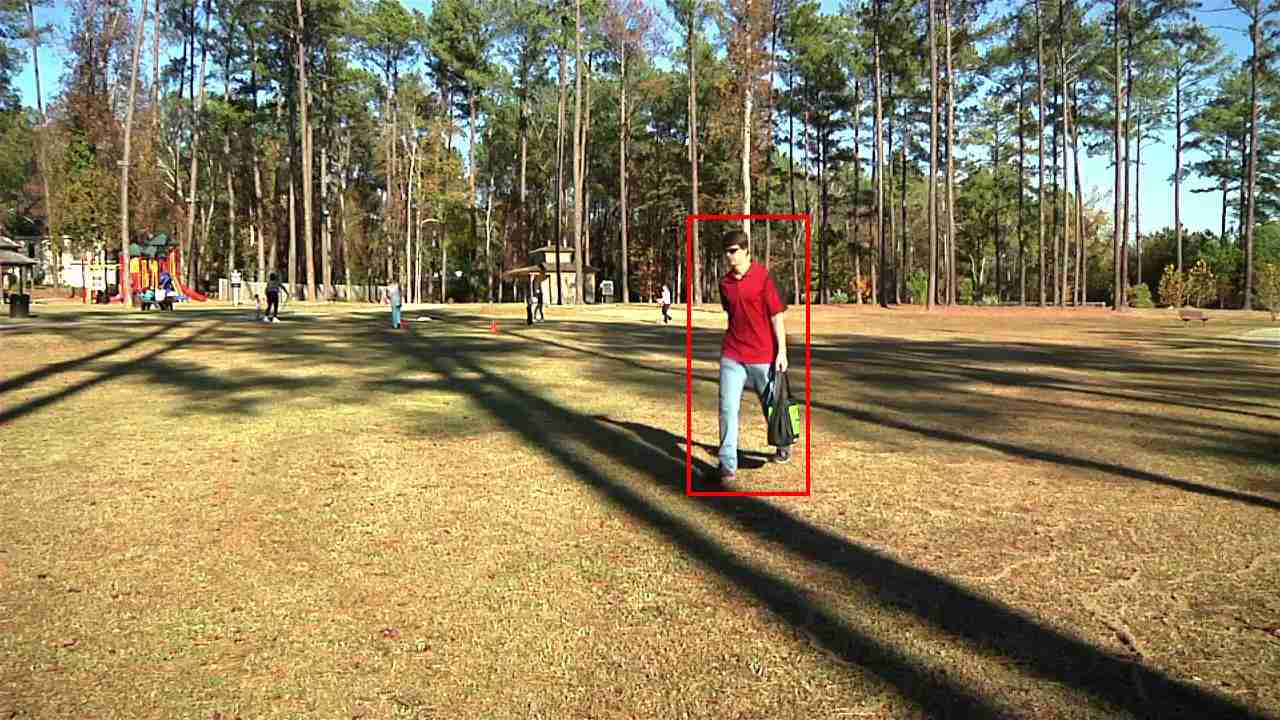}\\
      \includegraphics[width=0.32\textwidth]{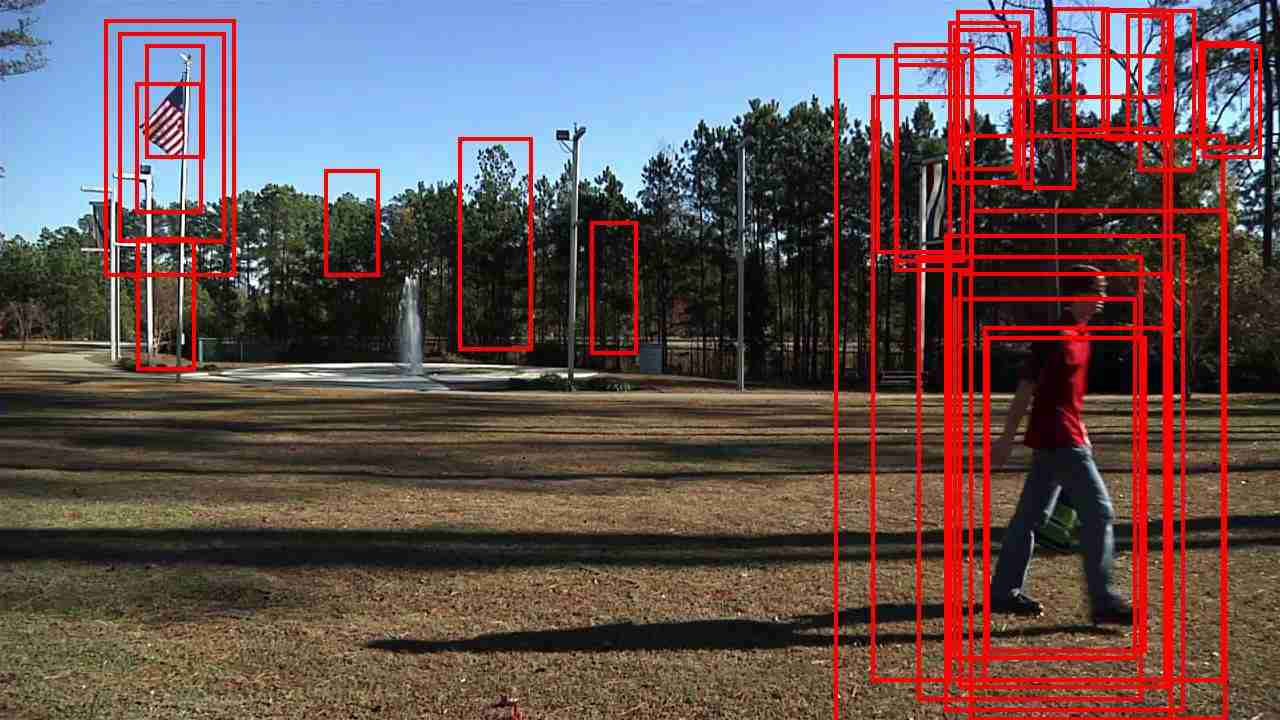}&
      \includegraphics[width=0.32\textwidth]{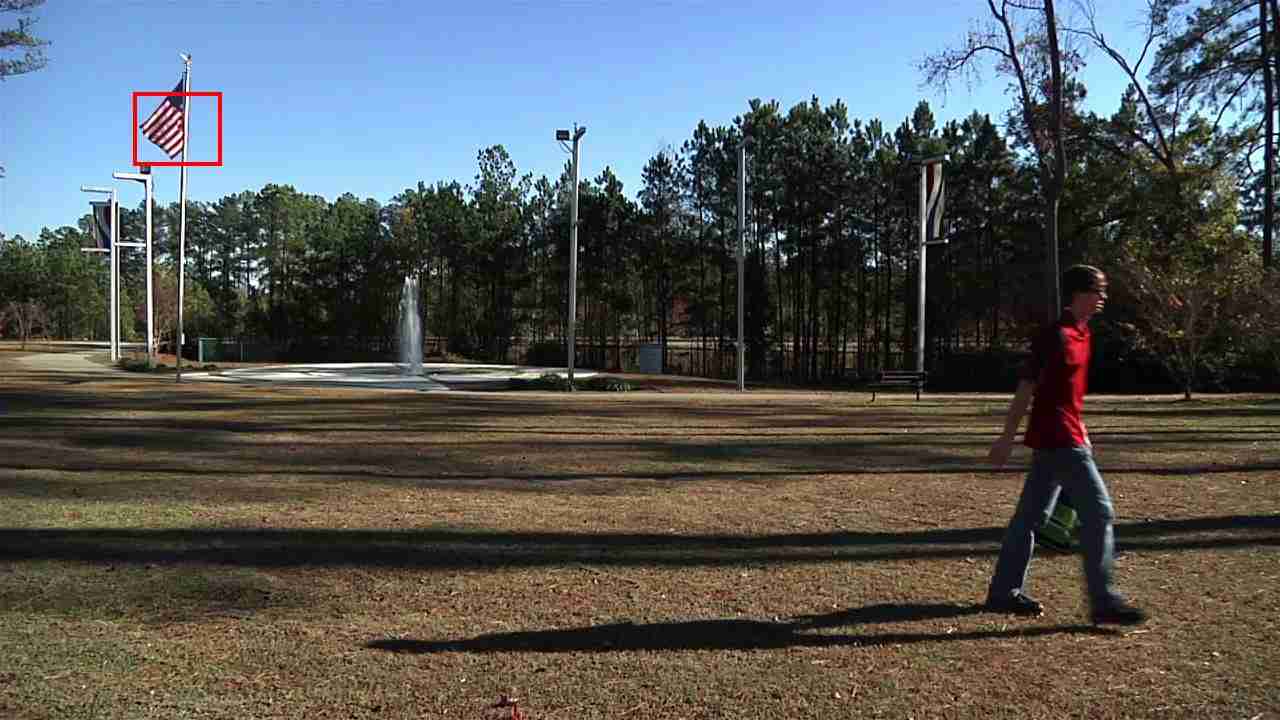}&
      \includegraphics[width=0.32\textwidth]{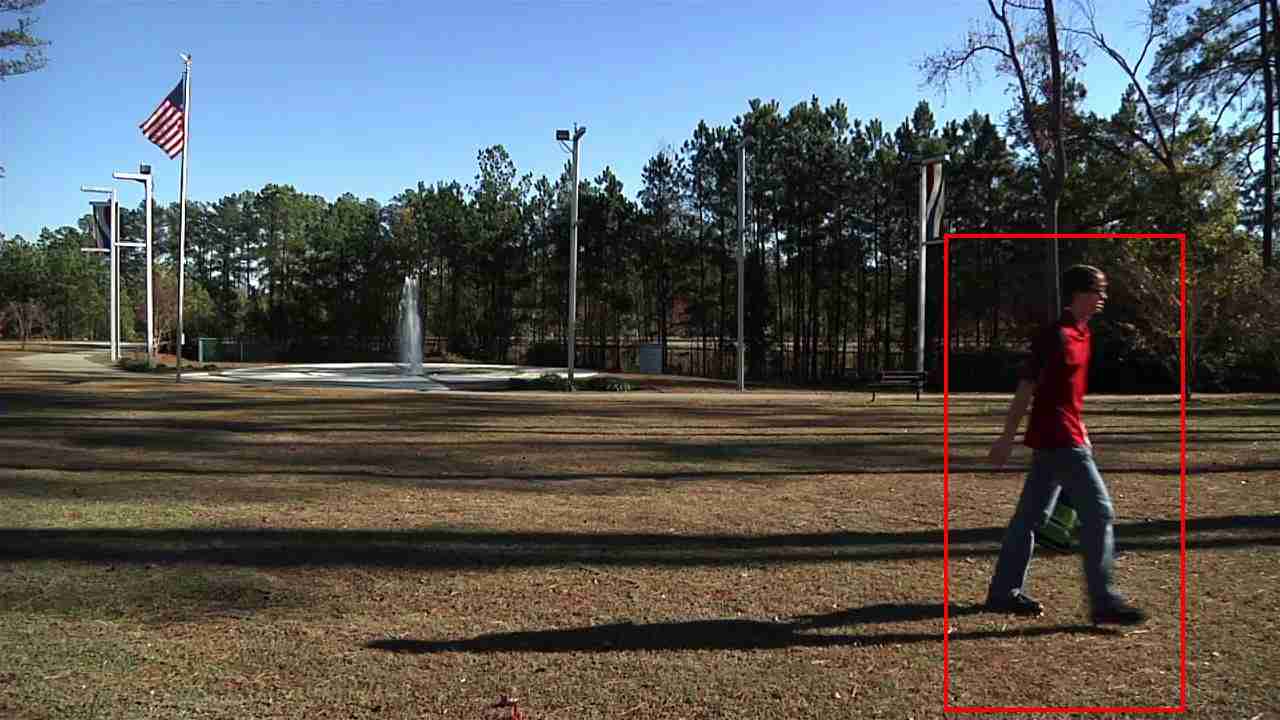}\\
      (a)&(b)&(c)
    \end{tabular}
  \end{center}
  \caption{Improved performance of simultaneous tracking and event recognition.
    (a)~Output of the Felzenszwalb \etal\ detector.
    (b)~Tracks produced by detection-based-tracking.
    (c)~Tracks produced by simultaneous tracking and event recognition.}
  \label{fig:objectsandtrackingandevents}
\end{figure*}

\begin{figure*}
  \begin{center}
    \begin{tabular}{@{}c@{\hspace*{2pt}}c@{\hspace*{2pt}}c@{}}
      \includegraphics[width=0.32\textwidth]{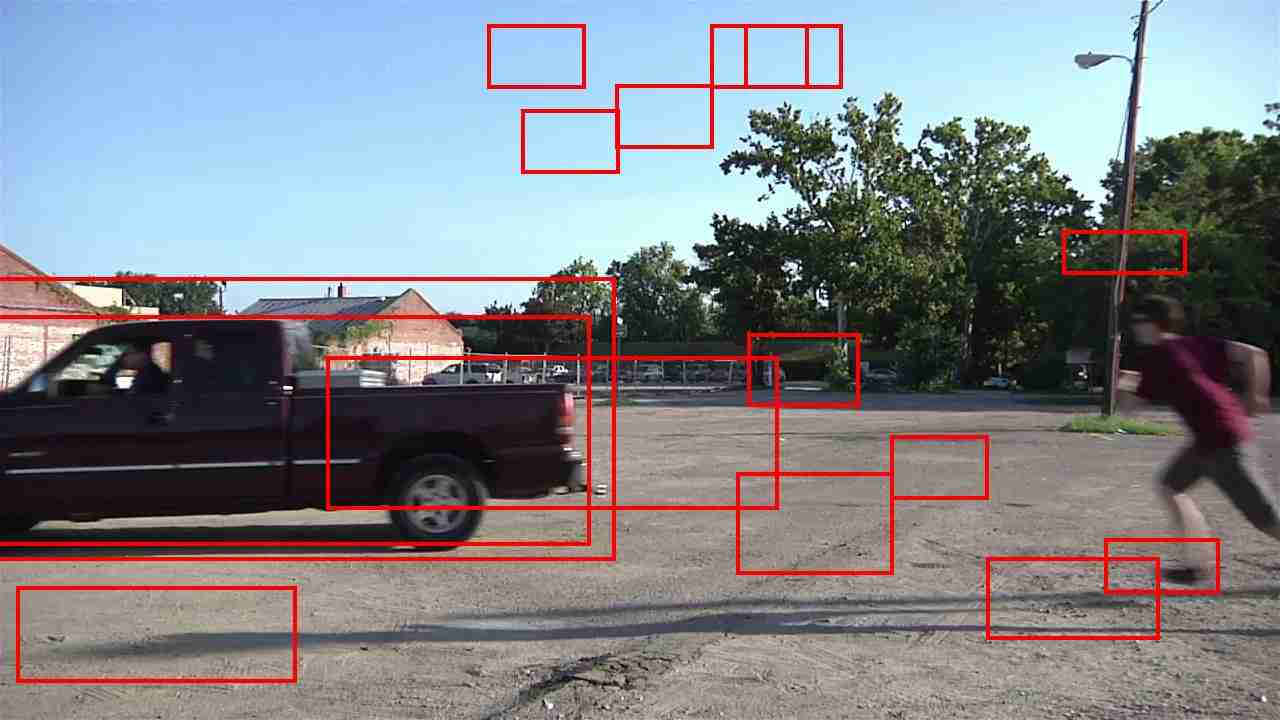}&
      \includegraphics[width=0.32\textwidth]{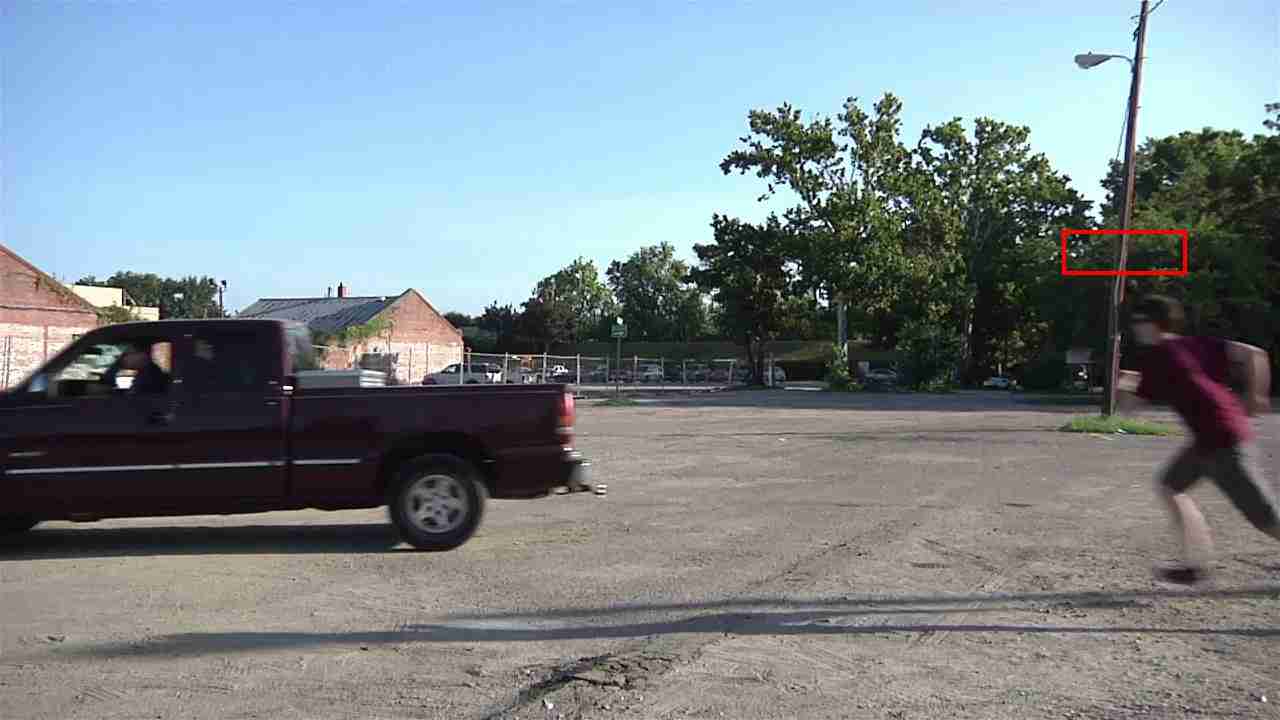}&
      \includegraphics[width=0.32\textwidth]{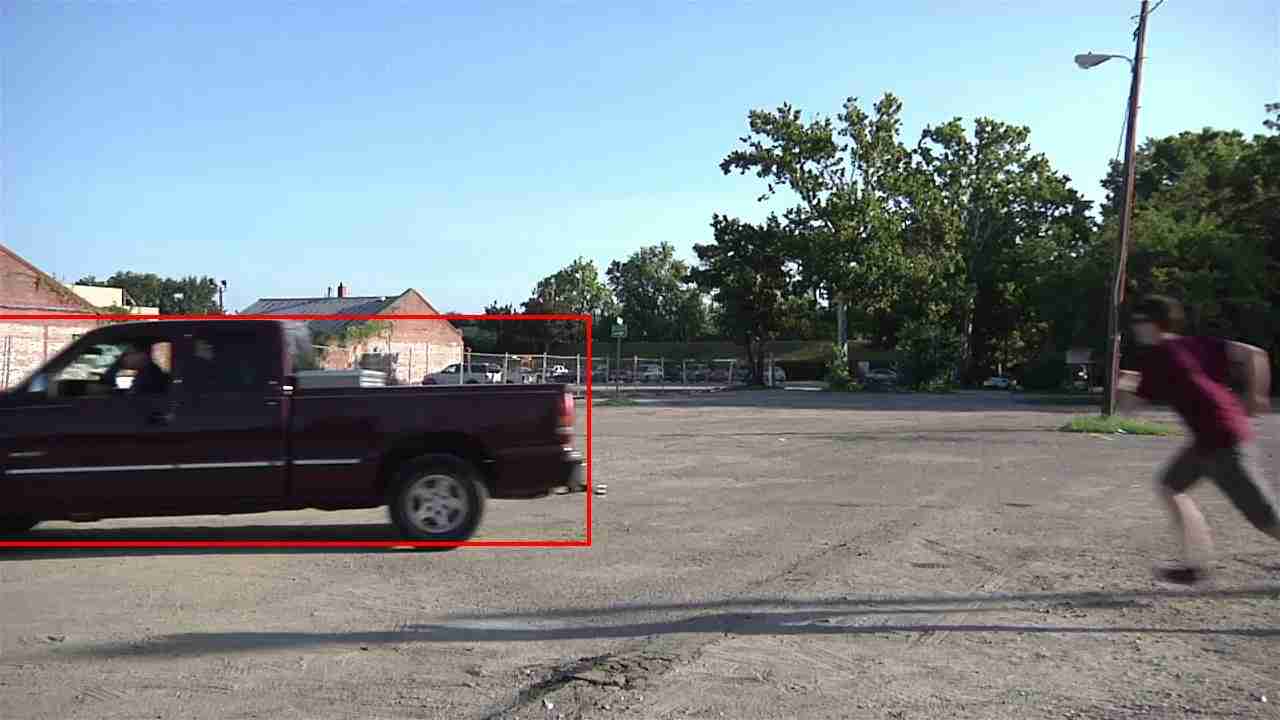}\\
      \includegraphics[width=0.32\textwidth]{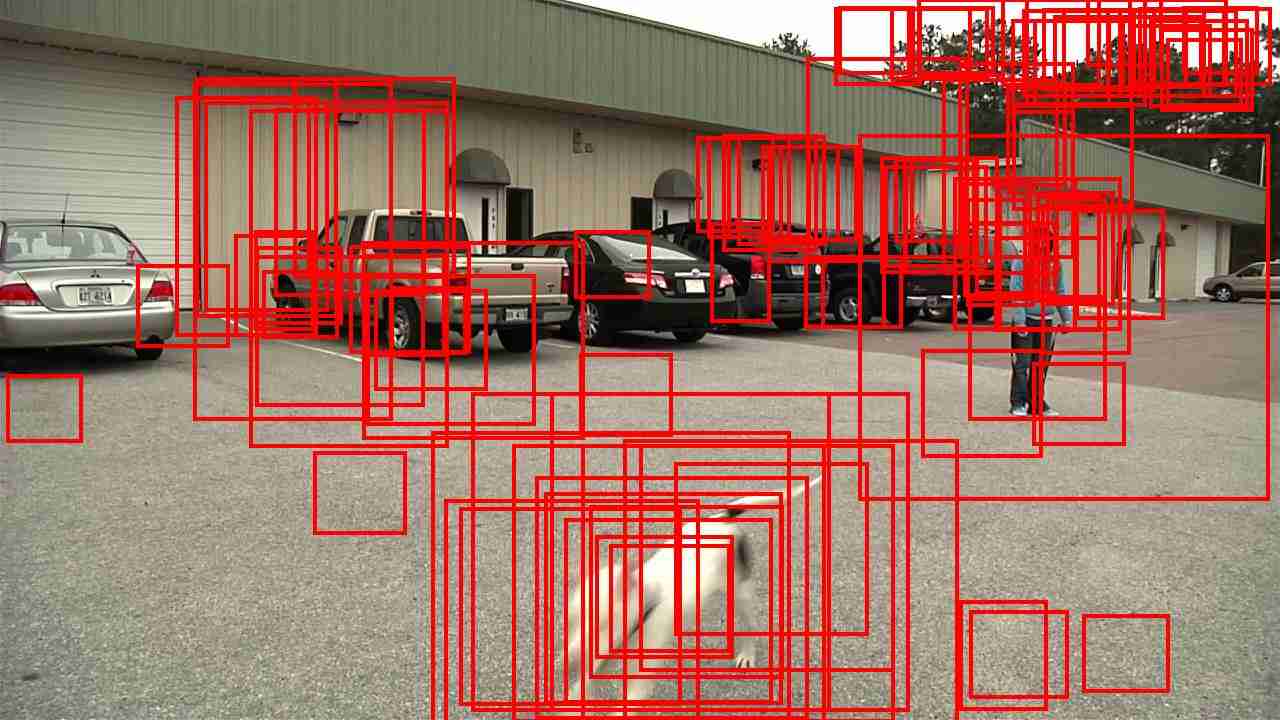}&
      \includegraphics[width=0.32\textwidth]{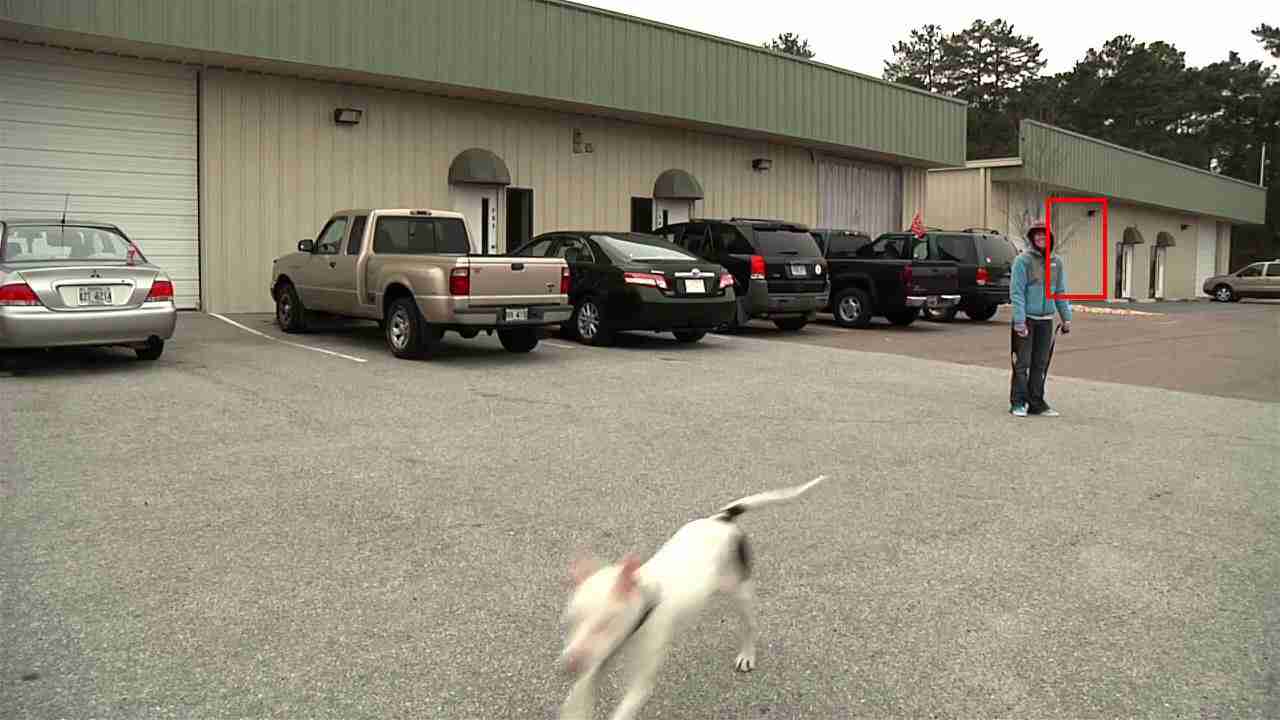}&
      \includegraphics[width=0.32\textwidth]{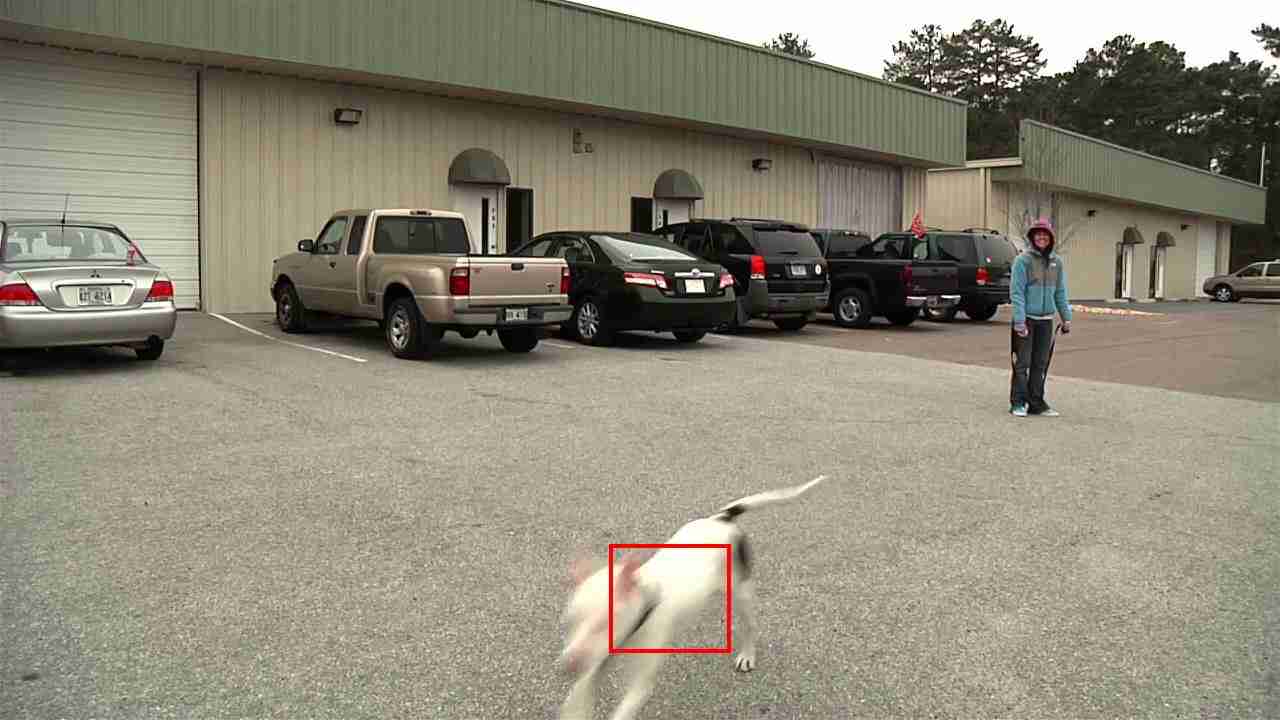}\\
      (a)&(b)&(c)
    \end{tabular}
  \end{center}
  \caption{Improved performance of simultaneous object detection, tracking, and
    event recognition.
    (a)~Output of the Felzenszwalb \etal\ detector.
    (b)~Tracks produced by detection-based-tracking.
    (c)~Tracks produced by simultaneous object-detection, tracking, and event
    recognition.}
  \label{fig:allthree}
\end{figure*}

Figure \ref{fig:objectsandtrackingandevents} demonstrates improved performance
of simultaneous tracking and event recognition~(c) over tracking~(b) in
isolation.
These results were obtained with object and event models that were trained
independently.\footnote{It would appear possible to co-train object and event
  models by combining Baum-Welch \citep{BaumPSW70, Baum72} with the training
  procedure for object models \citep{Felzenszwalb2010b}.}
The object models were trained on isolated frames using the standard
Felzenszwalb training software.
The event models were trained using tracks produced by the detection-based
tracking method described in Section~\ref{sec:detectionbasedtracking}.
It is difficult to track the person running with detection-based tracking alone
due to articulated appearance change and motion blur.
Imposing the prior of detecting \emph{running} biases the tracker to find the
desired track.

Figure \ref{fig:allthree} demonstrates improved performance
of simultaneous object detection, tracking, and event recognition~(c) over
object detection~(a) and tracking~(b) in isolation.
As before, these results were obtained with object and event models that were
trained independently.

\section{Conclusion}

Detection-base tracking using dynamic programming has a long history
\citep{Castanon1990, Wolf1989}, as do motion-profile-based approaches to event
recognition using HMMs \citep{Siskind1996, Starner98, Wang2009, Xu2002, Xu2005}.
Moreover, there have been attempts to integrate object detection and
tracking \citep{Li2008, Pirsiavash2011}, tracking and event recognition
\citep{Li2002}, and object detection and event recognition
\citep{Gupta2007,Moore1999,Peursum2005}.
However, we are unaware of prior work that integrates all three and does so in
a fashion that efficiently finds a global optimum to a simple unified cost
function.

We have demonstrated a general framework for simultaneous object detection,
tracking, and event recognition
Many object detectors can naturally be transformed into trackers by introducing
time into their cost functions, thus tracking every possible detection in each
frame.
Furthermore, the distance transform can be used to reduce the complexity of
doing so from quadratic to linear.
The common internal structure and algorithmic organization of object
detection, detection-based tracking, and event recognition further allows an
HMM-based approach to event recognition to be incorporated into the general
dynamic-programming approach.
This facilitates multidirectional information flow where not only can object
detection influence tracking and, in turn, event recognition, event
recognition can influence tracking and, in turn object detection.

\section*{Acknowledgments}
  \par\vspace*{-2ex}
This work was supported, in part, by NSF grant CCF-0438806, by the
Naval Research Laboratory under Contract Number N00173-10-1-G023, by the Army
Research Laboratory accomplished under Cooperative Agreement Number
W911NF-10-2-0060, and by computational resources provided by Information
Technology at Purdue through its Rosen Center for Advanced Computing.
Any views, opinions, findings, conclusions, or recommendations contained or
expressed in this document or material are those of the author(s) and do not
necessarily reflect or represent the views or official policies, either
expressed or implied, of NSF, the Naval Research Laboratory, the Office of
Naval Research, the Army Research Laboratory, or the U.S. Government.
The U.S. Government is authorized to reproduce and distribute reprints for
Government purposes, notwithstanding any copyright notation herein.

\bibliographystyle{plainnat}
\begin{small}
\bibliography{arxiv2012a}

\begin{thebibliography}{33}
\providecommand{\natexlab}[1]{#1}
\providecommand{\url}[1]{\texttt{#1}}
\expandafter\ifx\csname urlstyle\endcsname\relax
  \providecommand{\doi}[1]{doi: #1}\else
  \providecommand{\doi}{doi: \begingroup \urlstyle{rm}\Url}\fi

\bibitem[Baum(1972)]{Baum72}
L.~E. Baum.
\newblock An inequality and associated maximization technique in statistical
  estimation of probabilistic functions of a {M}arkov process.
\newblock \emph{Inequalities}, 3:\penalty0 1--8, 1972.

\bibitem[Baum and Petrie(1966)]{Baum1966}
L.~E. Baum and T.~Petrie.
\newblock Statistical inference for probabilistic functions of finite state
  {M}arkov chains.
\newblock \emph{Ann. Math. Stat}, 37:\penalty0 1554--63, 1966.

\bibitem[Baum et~al.(1970)Baum, Petrie, Soules, and Weiss]{BaumPSW70}
L.~E. Baum, T.~Petrie, G.~Soules, and N.~Weiss.
\newblock A maximization technique occuring in the statistical analysis of
  probabilistic functions of {M}arkov chains.
\newblock \emph{The Annals of Mathematical Statistics}, 41\penalty0
  (1):\penalty0 164--71, 1970.

\bibitem[Bui(2011)]{sri2011}
H.~Bui.
\newblock personal communication, 2011.

\bibitem[Castanon(1990)]{Castanon1990}
D.A. Castanon.
\newblock Efficient algorithms for finding the {K} best paths through a
  trellis.
\newblock \emph{IEEE Transactions on Aerospace and Electronic Systems},
  26\penalty0 (2):\penalty0 405--10, March 1990.

\bibitem[Comaniciu et~al.(2003)Comaniciu, Ramesh, and Meer]{Comaniciu2003}
D.~Comaniciu, V.~Ramesh, and P.~Meer.
\newblock Kernel-based object tracking.
\newblock \emph{{IEEE} Transactions on Pattern Analysis and Machine
  Intelligence}, 25\penalty0 (5):\penalty0 564--75, 2003.

\bibitem[Corso(2011)]{buffalo2011}
J.~Corso, 2011.
\newblock URL
  \url{http://www.cse.buffalo.edu/~jcorso/bigshare/mindseye_human_annotation_may11_buffalo.tar.bz}.

\bibitem[Everingham et~al.(2010)Everingham, Van~Gool, Williams, Winn, and
  Zisserman]{Everingham10}
M.~Everingham, L.~Van~Gool, C.~K.~I. Williams, J.~Winn, and A.~Zisserman.
\newblock The {PASCAL Visual Object Classes (VOC)} challenge.
\newblock \emph{International Journal of Computer Vision}, 88\penalty0
  (2):\penalty0 303--38, 2010.

\bibitem[Felzenszwalb and Huttenlocher(2004)]{Felzenszwalb2004}
P.~F. Felzenszwalb and D.~P. Huttenlocher.
\newblock Distance transforms of sampled functions.
\newblock Technical Report TR2004-1963, Cornell Computing and Information
  Science, 2004.

\bibitem[Felzenszwalb et~al.(2010{\natexlab{a}})Felzenszwalb, Girshick, and
  McAllester]{Felzenszwalb2010b}
P.~F. Felzenszwalb, R.~B. Girshick, and D.~McAllester.
\newblock Cascade object detection with deformable part models.
\newblock In \emph{Proceedings of the IEEE Computer Society Conference on
  Computer Vision and Pattern Recognition}, 2010{\natexlab{a}}.

\bibitem[Felzenszwalb et~al.(2010{\natexlab{b}})Felzenszwalb, Girshick,
  McAllester, and Ramanan]{Felzenszwalb2010a}
P.~F. Felzenszwalb, R.~B. Girshick, D.~McAllester, and D.~Ramanan.
\newblock Object detection with discriminatively trained part based models.
\newblock \emph{{IEEE} Transactions on Pattern Analysis and Machine
  Intelligence}, 32\penalty0 (9), September 2010{\natexlab{b}}.

\bibitem[Felzenszwalb et~al.(2003)Felzenszwalb, Huttenlocher, and
  Kleinberg]{FelzenszwalbHK03}
Pedro~F. Felzenszwalb, Daniel~P. Huttenlocher, and Jon~M. Kleinberg.
\newblock Fast algorithms for large-state-space {HMMs} with applications to web
  usage analysis.
\newblock In \emph{Neural Information Processing Systems}, 2003.

\bibitem[Freeman and Roth(1995)]{freeman1995}
W.~T. Freeman and M.~Roth.
\newblock Orientation histograms for hand gesture recognition.
\newblock In \emph{International Workshop on Automatic Face and Gesture
  Recognition}, pages 296--301, June 1995.

\bibitem[Gupta and Davis(2007)]{Gupta2007}
A.~Gupta and L.~S. Davis.
\newblock Objects in action: an approach for combining action understanding and
  object perception.
\newblock In \emph{Proceedings of the IEEE Computer Society Conference on
  Computer Vision and Pattern Recognition}, 2007.

\bibitem[Li and Chellappa(2002)]{Li2002}
Baoxin Li and Rama Chellappa.
\newblock A generic approach to simultaneous tracking and verification in
  video.
\newblock \emph{{IEEE} Transactions on Image Processing}, 11\penalty0
  (5):\penalty0 530--44, 2002.

\bibitem[Li and Nevatia(2008)]{Li2008}
Yuan Li and Ramakant Nevatia.
\newblock Key object driven multi-category object recognition, localization,
  and tracking using spatio-temporal context.
\newblock In \emph{Proceedings of the European Conference on Computer Vision},
  volume~IV, pages 409--22, 2008.

\bibitem[Moore et~al.(1999)Moore, Essa, and Heyes]{Moore1999}
D.~J. Moore, I.~A. Essa, and M.~H. Heyes.
\newblock Exploting human actions and object context for recognition tasks.
\newblock In \emph{Proceedings of the {$7^{\mbox{\scriptsize\em th}}$}
  International Conference on Computer Vision}, 1999.

\bibitem[Navatia(2011)]{usc2011}
R.~Navatia.
\newblock personal communication, 2011.

\bibitem[Otsu(1979)]{Otsu1979}
N.~Otsu.
\newblock A threshold selection method from gray-level histograms.
\newblock \emph{{IEEE} Transactions on Systems, Man and Cybernetics},
  9\penalty0 (1):\penalty0 62--6, January 1979.
\newblock ISSN 0018-9472.

\bibitem[Peursum et~al.(2005)Peursum, West, and Venkatesh]{Peursum2005}
P.~Peursum, G.~West, and S.~Venkatesh.
\newblock Combining image regions and human activity for indirect object
  recognition in indoor wide-angle views.
\newblock In \emph{Proceedings of the {$10^{\mbox{\scriptsize\em th}}$}
  International Conference on Computer Vision}, 2005.

\bibitem[Pirsiavash et~al.(2011)Pirsiavash, Ramanan, and
  Fowlkes]{Pirsiavash2011}
H.~Pirsiavash, D.~Ramanan, and C.~C. Fowlkes.
\newblock Globally-optimal greedy algorithms for tracking a variable number of
  objects.
\newblock In \emph{Proceedings of the IEEE Computer Society Conference on
  Computer Vision and Pattern Recognition}, pages 1201--8, 2011.

\bibitem[Saenko(2011)]{berkeley2011}
K.~Saenko, 2011.
\newblock URL
  \url{https://s3.amazonaws.com/Annotations/vaticlabels_C-D1_0819.tar.gz}.

\bibitem[Sala et~al.(2010)Sala, Macrini, and Dickinson]{Sala2010}
Pablo Sala, Diego Macrini, and Sven~J. Dickinson.
\newblock Spatiotemporal contour grouping using abstract part models.
\newblock In \emph{Proceedings of the {$10^{\mbox{\scriptsize\em th}}$} Asian
  Conference on Computer Vision}, volume~4, pages 539--52, 2010.

\bibitem[Shi and Tomasi(1994)]{shi1994}
J.~Shi and C.~Tomasi.
\newblock Good features to track.
\newblock In \emph{Proceedings of the IEEE Computer Society Conference on
  Computer Vision and Pattern Recognition}, pages 593--600, 1994.

\bibitem[Siskind and Morris(1996)]{Siskind1996}
J.~M. Siskind and Q.~Morris.
\newblock A maximum-likelihood approach to visual event classification.
\newblock In \emph{Proceedings of the Fourth European Conference on Computer
  Vision}, pages 347--60, April 1996.

\bibitem[Starner et~al.(1998)Starner, Weaver, and Pentland]{Starner98}
Thad Starner, Joshua Weaver, and Alex Pentland.
\newblock Real-time {A}merican sign language recognition using desk and
  wearable computer based video.
\newblock \emph{{IEEE} Transactions on Pattern Analysis and Machine
  Intelligence}, 20\penalty0 (12):\penalty0 1371--5, 1998.

\bibitem[Tomasi and Kanade(1991)]{tomasi1991}
C.~Tomasi and T.~Kanade.
\newblock Detection and tracking of point features.
\newblock Technical Report CMU-CS-91-132, Carnegie Mellon University, 1991.

\bibitem[Viterbi(1971)]{Viterbi1971}
A.~J. Viterbi.
\newblock Convolutional codes and their performance in communication systems.
\newblock \emph{{IEEE} Transactions on Communication}, 19:\penalty0 751--72,
  October 1971.

\bibitem[Wang et~al.(2009)Wang, Kuruoglu, Yang, Xu, and Yu]{Wang2009}
Zhaowen Wang, Ercan~E. Kuruoglu, Xiaokang Yang, Yi~Xu, and Songyu Yu.
\newblock Event recognition with time varying hidden {M}arkov model.
\newblock In \emph{Proceedings of the International Conference on Accoustic and
  Speech Signal Processing}, pages 1761--4, 2009.

\bibitem[Wolf et~al.(1989)Wolf, Viterbi, and Dixon]{Wolf1989}
J.~K. Wolf, A.M. Viterbi, and G.~S. Dixon.
\newblock Finding the best set of {K} paths through a trellis with application
  to multitarget tracking.
\newblock \emph{IEEE Transactions on Aerospace and Electronic Systems},
  25\penalty0 (2):\penalty0 287--96, March 1989.

\bibitem[Xu et~al.(2002)Xu, Ma, Zhang, and Yang]{Xu2002}
Gu~Xu, Yu-Fei Ma, HongJiang Zhang, and Shiqiang Yang.
\newblock Motion based event recognition using {HMM}.
\newblock In \emph{Proceedings of the International Conference on Pattern
  Recognition}, volume~2, 2002.

\bibitem[Xu et~al.(2005)Xu, Ma, Zhang, and Yang]{Xu2005}
Gu~Xu, Yu-Fei Ma, HongJiang Zhang, and Shi-Qiang Yang.
\newblock An {HMM}-based framework for video semantic analysis.
\newblock \emph{IEEE Trans. Circuits Syst. Video Techn.}, 15\penalty0
  (11):\penalty0 1422--33, 2005.

\bibitem[Yilmaz et~al.(2006)Yilmaz, Javed, and Shah]{Yilmaz2006}
A.~Yilmaz, O.~Javed, and M.~Shah.
\newblock Object tracking: A survey.
\newblock \emph{ACM Computing Surveys}, 38\penalty0 (4), December 2006.

\end{thebibliography}
\end{small}
\end{document}